\documentclass[10pt,journal,compsoc]{IEEEtran}

\def\IEEEcompsocdiamondline{}

\usepackage{hyperref}       % hyperlinks
\usepackage{url}            % simple URL typesetting
\usepackage{booktabs}       % professional-quality tables
\usepackage{nicefrac}       % compact symbols for 1/2, etc.
\usepackage{microtype}      % microtypography
\usepackage{xcolor}         % colors

\usepackage{amsmath, amssymb, amsfonts, amsthm, mathtools}
\usepackage[pdftex]{graphicx}

\newtheorem{theorem}{Theorem}
\newtheorem{lemma}{Lemma}
\newtheorem{corollary}{Corollary}
\newcommand{\centered}[1]{\overline{#1}}
\newcommand{\one}{\mathbf{1}}
\newcommand{\comment}[1]{}
\newcommand{\demo}[1]{\noindent{\em #1}}

\newcommand{\boxenddemo}{$\Box$}
\newcommand\numberthis{\addtocounter{equation}{1}\tag{\theequation}}
\ifCLASSOPTIONcompsoc
  \usepackage[nocompress]{cite}
\else
  \usepackage{cite}
\fi

\begin{document}

\title{Gaussian RBF Centered Kernel Alignment (CKA) in the Large Bandwidth Limit}

\author{Sergio A.\ Alvarez, alvarez@bc.edu\\\small
\vspace{0.1cm}
Department of Computer Science\\
Boston College\\
Chestnut Hill, MA 02467 USA	% }% <-this % stops a space
%\IEEEcompsocitemizethanks{\IEEEcompsocthanksitem S.\ A.\ Alvarez is with the
%Department of Computer Science, Boston College, Chestnut Hill, MA 02467 USA.\protect\\
%E-mail: alvarez@bc.edu
%}% <-this % stops an unwanted space
} % end author block

\markboth{}% Journal title in this space
{Alvarez\MakeLowercase{\textit{}}: Gaussian RBF CKA in the Large Bandwidth Limit}

\IEEEtitleabstractindextext{%
\begin{abstract}
We prove that Centered Kernel Alignment (CKA) based on a Gaussian RBF kernel converges to linear CKA in the large-bandwidth limit.
We show that convergence onset is sensitive to the geometry of the feature representations, and that representation eccentricity bounds the range of bandwidths for which Gaussian CKA behaves nonlinearly.
%Centered kernel alignment (CKA), also known as centered kernel-target alignment, is useful both as a similarity measure between kernels and as a kernel-based similarity measure between learned feature representations. We prove that, as a consequence of mean-centering of features, CKA similarity based on a Gaussian RBF kernel converges to linear kernel CKA in the large-bandwidth limit. We show that convergence onset is sensitive to the geometry of the feature representations, and that representation eccentricity bounds the range of bandwidths for which Gaussian CKA exhibits nonlinear behavior, which can aid in data-adaptive bandwidth selection. 
\end{abstract}

}

\maketitle

\IEEEdisplaynontitleabstractindextext

\IEEEraisesectionheading{\section{Introduction}}

Centered Kernel Alignment (CKA) was first proposed as a measure of similarity between kernels in the context of kernel learning~\cite{CortesEtAl2010, CortesCKA2012},
building on prior work on (non-centered) kernel-target alignment~\cite{CristianiniEtAl2001}.
Versions for functional data of CKA and the associated Hilbert-Schmidt Independence Criterion (HSIC)
were proposed later~\cite{FunctionalCKA2020}.
CKA has been used for multiple kernel learning (e.g.,~\cite{KernelClustering2014},~\cite{MultipleKernelLearning2017}).
%Such work often views CKA as a measure of similarity between kernels. 
CKA also enables measuring similarity between two feature representations of a set of data examples (e.g., two sets of neural network activation vectors), by comparing kernel-based similarity matrices (Gram matrices) of these representations~\cite{KornblithCKA2019}. Used in the latter manner, 
CKA and alternatives such as canonical correlation analysis (CCA) and orthogonal Procrustes distance~\cite{DingEtAlNeurIPS2021} can provide insight into the relationship between architectural features of a network such as width and depth, on one hand, 
and the network's learned representations~\cite{KornblithCKA2019, KornblithWideDeepCKAICLR2021}, on the other. CKA has been used to assess representation similarity in cross-lingual language models~\cite{CKAinNLP2020} and recurrent neural network dynamics~\cite{CKAforRNNSimilarity2019}, to identify potential drug side-effects~\cite{DrugSideEffects2019}, to differentiate among brain activity patterns~\cite{BrainActivityPatterns2017}, and to gauge the representational effects of watermarking~\cite{CKAforWatermarkEntanglement2021}, among others. % applications.

Any positive-definite symmetric kernel can be used as the base kernel for CKA, including linear, polynomial, and Gaussian radial basis function (RBF) kernels. Gaussian RBF kernels are of special interest due to their universality properties~\cite{KernelUniversality2011}, which can lead to
superior modeling. Understanding the behavior of CKA with Gaussian RBF kernels is, therefore, relevant to its use as a representational similarity metric.
Some prior work~\cite{KornblithCKA2019} reports finding little difference empirically between CKA similarity values based on Gaussian RBF and linear kernels.
In contrast, hyperparameter tuning experiments for~\cite{SarunPhDThesis2021} suggest that a noticeable difference can occur between CKA values for Gaussian RBF and linear kernels, but also that the difference becomes negligible as Gaussian bandwidth grows. 
We establish the latter phenomenon theoretically in the present paper, proving that Gaussian CKA converges to linear CKA as bandwidth approaches infinity, for all representations. We describe the convergence rate, as well. 
%, and provide a heuristic upper bound on the range of bandwidths for which Gaussian CKA behaves nonlinearly. 

We are not aware of previously published results on the large-bandwidth asymptotics of Gaussian CKA. 
A limit result for Gaussian kernel SVM classification appeared in~\cite{KeerthiLinGaussianSVMLimits2003}, but relies on an analysis of the dual formulation of the maximum-margin optimization problem that does not translate directly to kernel CKA. 

While we find that approximation of Gaussian RBF CKA by linear CKA for large bandwidths becomes apparent by elementary means, our proof shows that the result relies on centering of the feature maps.
Indeed, we show that a similar result does not hold for the analogous non-centered kernel-target alignment measure of~\cite{CristianiniEtAl2001}.
We also show that a geometric measure of representation eccentricity serves to bound the range of bandwidths for which Gaussian and linear CKA differ for a given representation. This can be helpful in data-adaptive Gaussian bandwidth selection. % for Gaussian RBF kernel CKA.
%, especially for lower-dimensional representations. % similarity.

\section{Background and notation}
\label{section:notation}

Our perspective is that of measuring similarity between two feature representations of the same set of data examples. 
%We generally follow the notation of~\cite{KornblithCKA2019}, as in Eq.~\ref{eq:HSIC} and Eq.~\ref{eq:CKA}. Also see~\cite{CortesCKA2012}. 
We briefly review the basic ingredients and notation related to kernel similarity and CKA below
(see~\cite{KornblithCKA2019},~\cite{CortesCKA2012}).

\subsection{Feature representations}
$X \in \mathbb R^{N \times p}$ and $Y \in \mathbb R^{N \times q}$ will denote matrices of $p$-dimensional
(resp., $q$-dimensional) feature vectors for the same set of $N$ data examples. Each row of $X$ or $Y$ consists of the feature vector for one of these examples. 
%Likewise, we let denote a matrix of $q$-dimensional feature vectors for the same set of examples (e.g., activations from a different network layer or network). 
We use subindices to indicate the rows of $X$ and $Y$ (e.g., $x_i$, $y_i$).
%: $x_i$ denotes the feature vector in the $i$-th row of $X$, and similarly for $y$.
%Our focus is on comparing the representations $X$, $Y$ by using kernel CKA. 

\subsection{Kernel similarity}

%The general outline of the CKA approach is visible in the linear case. 
%If the rows of matrices $X$, $Y$ are feature encodings of the same data examples by two alternative embeddings of the data into Euclidean space, then 
Linear similarity between feature encodings $X$, $Y$ 
can be expressed as similarity of their ``similarity structures'' 
(within-encoding dot product matrices) $X X^T$, $Y Y^T$~\cite{KornblithCKA2019}:
\begin{equation}
\|Y^T X\|^2_F = \frac{}{}\ tr \left ( X X^T Y Y^T \right ),
\label{eq:featureSimilarityEqualsGramSimilarity}
\end{equation}
where $\|\ \|_F$ is the Frobenius norm and $tr$ is the trace function. The left-hand side reflects similarity of the feature representations; the right-hand side reflects similarity of their respective self-similarity matrices. 

CKA (defined in section~\ref{subsec:HSICandCKA}) corresponds to a normalized version of Eq.~\ref{eq:featureSimilarityEqualsGramSimilarity}, extended to kernel similarity by replacing $X X^T$ and $Y Y^T$ by Gram matrices $\centered{K}(X) =
(\centered{k}(x_i,x_j))_{i,j}$ and $\centered{L}(Y) = (\centered{l}(y_i,y_j))_{i,j}$ for positive-definite symmetric kernel functions, $k$ and $l$. The bars indicate that columns have been mean-centered~\cite{GrettonEtAl2005},~\cite{CortesCKA2012}; see section~\ref{subsection:meanCentering}. 

Gram matrix entries can be viewed as inner products of the embedded images of the data examples in a high-dimensional reproducing kernel Hilbert space (RKHS)~\cite{HofmannKernelsML2008}. Thus, kernels allow modeling aspects of representations %that are 
not easily accessible to the linear version.
 
\subsection{HSIC and CKA}
\label{subsec:HSICandCKA}

The kernel similarity perspective applied to Eq.~\ref{eq:featureSimilarityEqualsGramSimilarity}
yields the Hilbert-Schmidt Independence Criterion (HSIC)~\cite{GrettonEtAl2005} shown in Eq.~\ref{eq:HSIC}, where $N$ denotes the number of data
examples, and the dependence on $X$ and $Y$ has been hidden for economy of notation.
Mean-centering of the Gram matrices is assumed to have been carried out as described
in section~\ref{subsection:meanCentering}, below. 
\begin{equation}
\text{HSIC}(K,L) = \frac{1}{(N-1)^2}\ tr\left ( \centered{K} \; \centered{L} \right )
\label{eq:HSIC}
\end{equation}
CKA (Eq.~\ref{eq:CKA}) is a normalized version of HSIC; it takes values in the interval $[0, 1]$, as observed in~\cite{CortesCKA2012}.
\begin{equation}
\text{CKA}(K,L) = \frac{\text{HSIC}(K,L)}{\sqrt{\text{HSIC}(K,K)\; \text{HSIC}(L,L)}}
\label{eq:CKA}
\end{equation}
%\cite{KornblithCKA2019} discusses advantages of CKA over alternatives for similarity-based comparison of feature representations, such as canonical correlation analysis (CCA) and its variants, and includes examples of the use of CKA to gain insight into learned representations in neural networks.
%We expand on the notation for kernels and Gram matrices below.

\subsection{Kernels and Gram matrices}
\label{subsection:kernelsAndGramMatrices}
We consider Gram matrices $K = \left ( k(x_i, x_j) \right )_{i,j}$ and $L = \left ( l(y_i, y_j) \right )_{i,j}$, where $k(u,v)$ and $l(u,v)$ denote positive-definite (p.d.) symmetric kernel functions. We focus on linear and Gaussian RBF kernels; the Euclidean (polynomial, degree $2$) kernel is also used. See Eq.~\ref{eq:kernelOptions}, where $\cdot$ is dot product and $|\ |$ is Euclidean norm. 
%We typically omit explicit references to feature matrices $X$, $Y$, except for emphasis, and where required in proofs. 
Gram matrices have size $N \times N$, regardless of the dimensionality of the feature representation.
\begin{align*}
&\text{Linear} &K_{\text{lin}} &= \left (x_i \cdot x_j \right )_{i,j} \\%\ \ \quad \qquad 
%&\text{ where } &\cdot \text{ is the usual dot product}\\
&\text{Gaussian, bandwidth } \sigma &K_{G(\sigma)} &= \left (e^\frac{-|x_i-x_j|^2}{2\sigma^2} \right )_{i,j} \\% \ \ \ \qquad\qquad 
%&\text{ where } &\sigma \text{ is the bandwidth} 
\numberthis \label{eq:kernelOptions}\\
&\text{Euclidean (quadratic)} &K_E(x,y) &= \left (|x_i-x_j|^{2} \right )_{i,j} \\ %\qquad 
%&\text{ where } &|\ | \text{ is the Euclidean norm}
\end{align*}
\normalsize
%The same subscripts as in Eq.~\ref{eq:kernelOptions} are used for the Gram matrices associated with the respective kernels: $K_{\text{lin}}$, $K_{G(\sigma)}$, and $K_E$ denote linear, Gaussian, and Euclidean Gram matrices, respectively, all based on the features in $X$; 

\paragraph*{\em Note on scaling of distances in Gaussian Gram matrices} Following the heuristic of~\cite{KornblithCKA2019}, we use $\sigma' = d_X \sigma$ instead of $\sigma$ in Eq.~\ref{eq:kernelOptions}
when computing the Gram matrix $K_{G(\sigma)}(X)$, where 
$d_X$ is the median distance between examples (rows) of $X$
(e.g., $\sigma = 2$ describes a bandwidth equal to twice the median distance between examples). Equivalently, we divide all inter-example distances $|x_i-x_j|$ by $d_X$. This ensures invariance of Gaussian HSIC under isotropic scaling~\cite{KornblithCKA2019}.

\subsection{Mean-centering}
\label{subsection:meanCentering}
Mean-centering the Gram matrices in Eq.~\ref{eq:HSIC} and Eq.~\ref{eq:CKA} is crucial both for kernel learning~\cite{CortesCKA2012} and for the results in this paper. Mean-centering of the Gram matrices corresponds to centering the embedded features in the RKHS.
%, as one normally does when computing covariance or correlation. 
Centering in the embedding space appeared in early work on kernel PCA~\cite{ScholkopfSmolaMullerKernelPCA1998}.
%indeed, centering originally appeared in the context in the form of a cross-covariance operator~\cite{FukumizuEtalCrossCovariance2004}.

We default to column mean-centering, that is, we ensure that each column has mean zero. Thus, we multiply by the centering matrix, $H$, on the left as in Eq.~\ref{eq:columnCentering} 
(where $I_N$ is the $N \times N$ identity matrix and $\one \one^T$ is an $N \times N$ matrix of ones), 
with individual entries as in Eq.~\ref{eq:columnCenteringEntries}. 
For row mean-centering, $H$ would multiply from the right and the $k$ summation would range over columns instead of rows. 
\begin{subequations}
\begin{align}
&\centered{K} = H K, \text{ where } H = I_N - \frac{1}{N} \one \one^T\label{eq:columnCentering}\\
&\centered{K}_{i,j} = K_{i,j} - \frac{1}{N} \sum_{k=1}^N K_{k,j}\label{eq:columnCenteringEntries}
\end{align}
\label{eq:columnCenteringSet}
\end{subequations}
%If column mean-centering is carried out as in Eq.~\ref{eq:columnCentering}, then the entry in the $i$-th row and $j$-th column of the %centered matrix will be as in Eq.~\ref{eq:columnCenteringEntries}.
%It is also possible to perform successive mean-centering of both rows and columns, as in~\cite{CortesCKA2012}. Doing so would add a third %term to the right-hand side of Eq.~\ref{eq:columnCenteringEntries} that involves a double summation with a $1/N^2$ factor.
Our results hold for mean-centering of either rows or columns, and for simultaneous centering of both. The latter case~\cite{CortesCKA2012} involves multiplication by $H$ on both sides in Eq.~\ref{eq:columnCentering}; two terms are added to the right in Eq.~\ref{eq:columnCenteringEntries}, corresponding to subtracting $1/N$ times the column-sum of the current Eq.~\ref{eq:columnCenteringEntries}. 
%involving a double summation over rows and columns, with $1/N^2$ as a factor.

\section{Convergence proof and derivation of a bandwidth-selection heuristic}

We prove convergence of Gaussian CKA to linear CKA for large bandwidths in
section~\ref{subsection:GaussianCKAConvergesToLinearCKA}, and provide a
data-dependent criterion that bounds the range of Gaussian
bandwidths for which nonlinear behavior occurs, in Section~\ref{subsection:dataDependentSigmaBound}.

\subsection{Linear CKA approximation of Gaussian RBF CKA}
\label{subsection:GaussianCKAConvergesToLinearCKA}
Our main result is Theorem~\ref{thm:limCKAGequalsCKAL}. We focus on the case
in which only one of the two Gram matrix parameters uses a Gaussian RBF kernel with bandwidth $\sigma \rightarrow \infty$, as doing so leads to cleaner proofs; the other, $L$, is assumed to be any fixed, positive-definite symmetric kernel. The result holds equally if both Gram matrices are Gaussian with bandwidths approaching infinity (see Appendix in Supplemental Material).

\begin{theorem}
$\text{CKA}(K_{G(\sigma)},L) = \text{CKA}(K_{\text{lin}},L) + O \left ( \frac{1}{\sigma^2} \right )$ as $\sigma \rightarrow \infty$.
In particular, $\text{CKA}(K_{G(\sigma)},L)$ %\xrightarrow[\sigma \to \infty] {} 
converges to $\text{CKA}(K_{\text{lin}},L)$ as $\sigma \rightarrow \infty$.
The result also holds if both kernels are Gaussian RBF kernels; in that case, the limit as both bandwidths approach infinity is $\text{CKA}(K_{\text{lin}}, L_{\text{lin}})$ 
(i.e., it is $\text{CKA}(K_{\text{lin}}(X), K_{\text{lin}}(Y))$).
\label{thm:limCKAGequalsCKAL}
\end{theorem}

We prove Theorem~\ref{thm:limCKAGequalsCKAL} by showing first, in Lemma~\ref{lemma:limCKAGequalsCKAE}, that Gaussian CKA converges to Euclidean CKA, and then, as a Corollary to Lemma~\ref{lemma:HSICEequalsNegative2HSICL}, that Euclidean
CKA and linear CKA are identical. 
%Notation is as in the Introduction and section~\ref{section:notation}.

\paragraph*{\em Note} Our results rely crucially on mean-centering of features in Eqs.~\ref{eq:HSIC},~\ref{eq:CKA}. Indeed, the analog of Theorem~\ref{thm:limCKAGequalsCKAL} does not hold if, as in~\cite{CristianiniEtAl2001}, features are not centered. % in Eqs.~\ref{eq:HSIC},~\ref{eq:CKA}. 
This is easy to see by a direct calculation for the example in which feature matrix $X$ is the $2 \times 2$ identity and feature matrix $Y$ is a $2 \times 2$ of ones, except for a single off-diagonal $0$. Linear CKA equals $3/\sqrt{14}$ in that case, while Gaussian CKA is $1$ for all $\sigma$, as $X$ and $Y$ have identical distance matrices after scaling by the median distances $d_X$, $d_Y$ as described in the note in section~\ref{subsection:kernelsAndGramMatrices}.
Details are provided in the Appendix.

\begin{lemma}
$\text{CKA}(K_{G(\sigma)},L) = \text{CKA}(K_E,L) + O \left ( \frac{1}{\sigma^2} \right )$ as $\sigma \rightarrow \infty$, for any positive-definite kernel, $L$.
\label{lemma:limCKAGequalsCKAE}
\end{lemma}

\demo{Proof of Lemma~\ref{lemma:limCKAGequalsCKAE}:}
Let $\alpha_{i,j} = |x_i-x_j|$. %and\ $\beta_{i,j} = |y_i-y_j|$
Also, let $\sigma_X = d_X\sigma$, %and\ $\sigma_Y = d_Y\sigma$. 
where $d_X$ is the median pairwise distance between feature vectors in $X$. %each of $X$ and $Y$.  
Then, by Eq.~\ref{eq:kernelOptions} and Eq.~\ref{eq:columnCenteringEntries}, the centered Gram matrix $\centered{K_{G(\sigma)}}$ has the entries shown in Eq.~\ref{eq:centeredGaussianGramMatrix}. Mean-centering is an indispensable ingredient, as we will see.
\begin{equation}
\centered{K_{G(\sigma)}}_{i,j} = e^{-\frac{\alpha^2_{i,j}}{2\sigma^2_X}} 
 - \frac{1}{N}\sum_{k=1}^N e^{-\frac{\alpha_{k,j}^2}{2\sigma^2_X}}
\label{eq:centeredGaussianGramMatrix}
\end{equation}
The argument uses the first-order series expansion of the exponential function at the origin (Eq.~\ref{eq:expPowerSeries}). 
%, together with mean-centering of the feature maps (Eq.~\ref{eq:centeredGaussianGramMatrix}). 
\begin{equation}
e^{-u} = 1 - u + O(u^2) \quad \text{ as } u \rightarrow 0
\label{eq:expPowerSeries}
\end{equation}
Applying Eq.~\ref{eq:expPowerSeries} to the exponential terms 
of Eq.~\ref{eq:centeredGaussianGramMatrix} as $\sigma \rightarrow \infty$, the importance of mean-centering becomes apparent. Due to subtraction of the sum for the column mean in Eq.~\ref{eq:centeredGaussianGramMatrix},
the constant $1$ terms in Eq.~\ref{eq:expPowerSeries} cancel, leaving only the %corresponding 
centered Euclidean Gram matrix entries and higher-order terms: 
$$
\centered{K_{G(\sigma)}}_{i,j} = - \frac{\alpha^2_{i,j}}{2\sigma^2_X} 
 + \frac{1}{N}\sum_{k=1}^N \frac{\alpha_{k,j}^2}{2\sigma^2_X} 
 + O \left ( \frac{1}{\sigma^4} \right )
$$
The form of the residual term persists in the HSIC expression of Eq.~\ref{eq:HSIC}:
$$
\aligned
&(N-1)^2\, \text{HSIC}(K,L) = tr \left ( \centered{K}_{G(\sigma)} \centered{L} \right )\\
%&= \sum_{i=1}^N \sum_{j=1}^N \left ( e^{-\frac{\alpha_{i,j}^2}{2\sigma^2_X}} - \frac{1}{N}
%\sum_{k=1}^N e^{-\frac{\alpha_{k,j}^2}{2\sigma^2_X}} \right ) \centered{L}_{j,i} \\
&= \sum_{i=1}^N \sum_{j=1}^N \left ( -\frac{\alpha_{i,j}^2}{2\sigma^2_X} + \frac{1}{N}\sum_{k=1}^N \frac{\alpha_{k,j}^2}{2\sigma^2_X} \right )  
\centered{L}_{j,i} \; + \; O\left( \frac{1}{\sigma^4} \right )\\
\endaligned
$$

Multiplying by $\sigma^2_X$, we obtain the connection in Eq.~\ref{eq:HSICErrorTerms}
between $\text{HSIC}(K_{G(\sigma)},L)$ and $\text{HSIC}(K_E,L)$:
\begin{equation}
\aligned
&\sigma^2_X (N-1)^2\, \text{HSIC}(K_{G(\sigma)},L)\\ 
&= \sum_{i=1}^N \sum_{j=1}^N \left ( -\alpha_{i,j}^2 + \frac{1}{N}\sum_{k=1}^N \alpha_{k,j}^2 \right )  
\centered{L}_{j,i} \; + \; O\left( \frac{1}{\sigma^2} \right )\\
&= (N-1)^2\, \text{HSIC}(K_E,L) \; + \; O\left( \frac{1}{\sigma^2} \right )
\endaligned
\label{eq:HSICErrorTerms}
\end{equation}
An identity analogous to Eq.~\ref{eq:HSICErrorTerms} relates $\text{HSIC}(K_{G(\sigma)},K_{G(\sigma)})$ %and $\text{HSIC}(L,L)$ 
to its Euclidean version. In the latter case, the higher-order error terms in the HSIC expression are $O(\frac{1}{\sigma^6})$ before multiplying by $\sigma^2_X$, because they result from a product of a $O(\frac{1}{\sigma^4})$ error term from one of the two Gaussian Gram matrices with the $O(\frac{1}{\sigma^2})$ leading term in the other factor; multiplication by 
$\sigma^4_X$ is required to obtain a $O(\frac{1}{\sigma^2})$ residual,
which will become a $\sigma^2_X$ factor after taking the square root in the CKA denominator (below).

Lemma~\ref{lemma:limCKAGequalsCKAE} follows by dividing the HSIC components described above
to form CKA as in Eq.~\ref{eq:CKA}:
\begin{equation}
\aligned
& \text{CKA}(K_{G(\sigma)},L) = \frac{\text{HSIC}(K_{G(\sigma)},L)}{\sqrt{\text{HSIC}(K_{G(\sigma)},K_{G(\sigma)}) \text{HSIC}(L,L)}}\\
&= \text{CKA}(K_E,L)\; +\; O \left (\frac{1}{\sigma^2} \right ) \quad\quad\quad\quad 
\endaligned
\label{eq:CKAErrorTerms}
\end{equation}
The case of two Gaussian kernels with bandwidths approaching $\infty$ (whether equal to one another or not) follows by considering one kernel argument at a time, freezing the bandwidth of the other,
and using the scalar triangle inequality. Details appear in the Appendix. \boxenddemo

\begin{lemma}
${\text{HSIC}}(K_E,L) = -2 {\text{HSIC}}(K_{\text{lin}},L)$, for any p.d.\ $L$.
\label{lemma:HSICEequalsNegative2HSICL}
\end{lemma}
\demo{Proof:}
First, we show that the column-centered Gram matrix entries $\centered{K_E}_{i,j}$ can be written as $-2 \centered{K_{\text{lin}}}_{i,j} + \delta_i$, where the $\delta_i$ term depends only on the row, $i$, not the column, $j$:
\small
$$
\aligned
&\centered{K_E}_{i,j} = |x_i-x_j|^2 - \frac{1}{N}\sum_{k=1}^N |x_k-x_j|^2\\
%&\centered{K_E}_{i,j} = |x_i-x_j|^2 - \frac{1}{N}\sum_{k=1}^N |x_k-x_j|^2\\
&= |x_i|^2 + |x_j|^2 - 2 x_i \cdot x_j - \frac{1}{N}\sum_{k=1}^N |x_k|^2 - \frac{1}{N}\sum_{k=1}^N |x_j|^2 + \frac{2}{N}\sum_{k=1}^N x_k \cdot x_j\\
&= - 2 \left ( x_i \cdot x_j - \frac{1}{N}\sum_{k=1}^N x_k \cdot x_j \right ) + |x_i|^2 - \frac{1}{N}\sum_{k=1}^N |x_k|^2\\
\endaligned
$$
\normalsize
The quantity in parentheses is $\centered{K_{\text{lin}}}_{i,j}$, and the remaining term 
$\delta_i = |x_i|^2 - \frac{1}{N}\sum_{k=1}^N |x_k|^2$ depends only on $i$, as stated. %the row index, $i$.
We can now relate the HSIC expressions:
$$
\aligned
&{\text{HSIC}}(K_E,L) = tr \left ( \centered{K_E} \, \centered{L} \right )\\
%&= \sum_{i=1}^N \sum_{j=1}^N \centered{K_E}_{i,j} \centered{L}_{j,i}\\
&= \sum_{i=1}^N \sum_{j=1}^N \left ( -2\centered{K_{\text{lin}}}_{i,j} + \delta_i \right ) \centered{L}_{j,i}\\
&= -2 \sum_{i=1}^N \sum_{j=1}^N \centered{K_{\text{lin}}}_{i,j} \centered{L}_{j,i}
+ \sum_{i=1}^N \sum_{j=1}^N \delta_i \centered{L}_{j,i}\\
&= -2 \text{HSIC}(K_{\text{lin}}, L) + \sum_{i=1}^N \delta_i \sum_{j=1}^N \centered{L}_{j,i}\\
%&= 4\sum_{i=1}^N \sum_{j=1}^N \centered{K_{\text{lin}}}_{i,j} \centered{L_{\text{lin}}}_{j,i}
%-2\sum_{i=1}^N \sum_{j=1}^N \centered{K_{\text{lin}}}_{i,j} \epsilon_j \\
%&- 2\sum_{i=1}^N \sum_{j=1}^N \delta_i \centered{L_{\text{lin}}}_{j,i} + \sum_{i=1}^N \sum_{j=1}%^N \delta_i \epsilon_j\\
\endaligned
$$
Since the columns of $\centered{L}$ have mean zero, the sum on the right is zero.
If row-centering is used instead of column-centering, the conclusion follows
by first expressing the entries of the Euclidean Gram matrix as $-2$ times the %corresponding
linear Gram matrix entries plus a term that depends only on the column.
\boxenddemo

\begin{corollary}
\label{corollary:CKAEequalsCKAlin}
$\text{CKA}(K_E, L) = \text{CKA}(K_{\text{lin}},L)$ for any kernel $L$.
\end{corollary}

Theorem~\ref{thm:limCKAGequalsCKAL} follows from Lemma~\ref{lemma:limCKAGequalsCKAE} and Corollary~\ref{corollary:CKAEequalsCKAlin}.

\subsection{Bounding the nonlinear Gaussian bandwidth range}
\label{subsection:dataDependentSigmaBound}
A heuristic bound on the $O(\frac{1}{\sigma^2})$ term in Theorem~\ref{thm:limCKAGequalsCKAL}
follows along the same lines as the proof of Lemma~\ref{lemma:limCKAGequalsCKAE},
by examining how the residual grows as the CKA terms are assembled, and how its magnitude is constrained by geometric characteristics of the feature representations. We consider this
issue in the case in which a Gaussian kernel is used for both feature maps, as this potentially
maximizes the residual due to contributions from both Gram matrix factors in Lemma~\ref{lemma:limCKAGequalsCKAE}.
We show, below, how convergence onset reflects the representation-specific range of variation of pairwise distances between data examples in feature space.

The power series for $e^{-u}$ in Eq.~\ref{eq:expPowerSeries} is alternating if $u > 0$, with decreasing terms after the second if $u \le 1$; the $O(u^2)$ error term in Eq.~\ref{eq:expPowerSeries} is then 
no larger than $u^2/2$. {\em Relative} error with respect to the linear $u$
term is $u/2$. Multiplying or dividing two terms roughly doubles relative error, since
$(1+a)^2 = 1 + 2a + O(a^2)$ and $(1+b)/(1-b) = (1+b)(1 + b + O(b^2)) = 1 + 2b + O(b^2)$; 
likewise, since $\sqrt{1+a} = 1 + \frac{a}{2} + O(a^2)$, taking square roots halves relative error.
 
Let $u = \max(\alpha^2_{i,j} / (2 \sigma^2_{X}), \beta^2_{i,j} / (2 \sigma^2_{Y}))$; then 
each of the exponential arguments $\frac{\alpha_{i,j}^2}{2\sigma^2_X}, \frac{\beta_{i,j}^2}{2\sigma^2_Y}$ in the Gram matrices in the proof of Lemma~\ref{lemma:limCKAGequalsCKAE} is bounded by $u$. It follows by the error considerations in the
preceding paragraph that HSIC relative error in Eq.~\ref{eq:HSICErrorTerms} is approximately bounded by $u$, and CKA relative error in Eq.~\ref{eq:CKAErrorTerms} %in Lemma~\ref{lemma:limCKAGequalsCKAE} 
is bounded by $2u$. Notice that $u \le \frac{\rho^2}{2\sigma^2}$, where the 
``representation eccentricity'', $\rho$, is defined as in Eq.~\ref{eq:rho}
(so $\rho$ depends only on pairwise distances, and $\rho >= 1$, with equality only if
all pairwise distances are equal). 
\begin{equation}
\rho = \max \left ( \frac{\text{diam}(X)}{d_X}, \frac{\text{diam}(Y)}{d_Y} \right )
\label{eq:rho}
\end{equation}
% and $\sigma_{XY} = \min(\sigma_X, \sigma_Y)$. 
Thus, CKA relative error is approximately bounded by $(\rho/\sigma)^2$, so that Gaussian RBF CKA approximately equals Euclidean CKA for bandwidths $\sigma \gg \rho$. 
For lower-dimensional representations $X$, $Y$, the representation eccentricity, $\rho$, 
provides a data-sensitive 
approximate threshold between nonlinear and linear regimes for Gaussian CKA. 
If feature dimensionality is high, concentration of Euclidean distance~\cite{AggarwalEtAlHighDDistance2001} will make $\rho \approx 1$; hence, for high-dimensional representations, one expects that Gaussian CKA will behave linearly if $\sigma \gg 1$. 

%We note in passing that robust versions of the eccentricity, $\rho$ (Eq.~\ref{eq:rho}), could also be considered, in terms of suitable pairs of quantiles of the distribution of pairwise distances. We will not pursue such alternative definitions in the present paper.

\section{Experimental illustration}
\label{section:experiments}
In this section we describe the results of a limited number of experiments that compare CKA similarity of neural feature representations based on Gaussian kernels of different bandwidths, with linear CKA similarity. 
For simplicity, we restrict attention to the case in which both of the kernels $K, L$ 
in Eq.~\ref{eq:CKA} are of the same type, either Gaussian RBF kernels of equal bandwidth, or standard linear kernels. Software for these experiments is available from the author upon request.

\paragraph*{\em Notation} 
In this section, we write $\text{CKA}_{G(\sigma)}$ as shorthand for $\text{CKA}(K_{G(\sigma)}, K_{G(\sigma)})$. Likewise, we abbreviate $\text{CKA}(K_{\text{lin}}, L_{\text{lin}})$ 
(i.e., $\text{CKA}(K_{\text{lin}}(X), K_{\text{lin}}(Y))$) as $\text{CKA}_{\text{lin}}$.

\subsection{Experimental setup}
\label{section:expSetup}

\paragraph*{\em Data sets}
We used sample OpenML data sets~\cite{OpenML2013} (CC BY 4.0 license, \url{https://creativecommons.org/licenses/by/4.0/}) for classification %~\cite{scikit-learn} 
(splice, tic-tac-toe, wdbc, optdigits, wine, dna) 
and regression (cpu, boston, Diabetes(scikit-learn), stock, balloon, cloud); for data sets with multiple versions, we used version 1.
These data sets were selected based on two considerations only: smaller size, in order to eliminate the need for GPU acceleration and reduce environmental impact; and an absence of missing values, to simplify the development of internal feature representations for the CKA computations. Data set
sizes are given in Table~\ref{table:datasetSizes}. 

\begin{table*}[h!]
\caption{Data set sizes.}	% 50 reps
\centering
\begin{tabular}{lccclcc}
\toprule
\multicolumn{3}{c}{Classification} &&\multicolumn{3}{c}{Regression}\\
data set &examples &attributes &{}&data set &examples &attributes\\
\cmidrule{1-3} \cmidrule(l){5-7}
splice &3190 &60 &{}&cpu &209 &7\\
tic-tac-toe &958 &9 &{}&boston &506 &13\\
wdbc &569 &30 &{}&Diabetes(scikit-learn) &442 &10\\
optdigits &5620 &64 &{}&stock &950 &9\\
wine &178 &13 &{}&balloon &2001 &1\\
dna &3186 &180 &{}&cloud &108 &5\\
\bottomrule
\end{tabular}
\normalsize
\label{table:datasetSizes}
\end{table*}

\paragraph*{\em Neural feature representations}
Fully-connected neural networks (NN) with two hidden layers were used. Feature encodings $X$, $Y$ were the sets of activation vectors of the first and second hidden layers, respectively. Alternative NN widths, $w = 16, 32, 64, 128, 256, 512, 1024$ were tested, with $w$ and $\frac{w}{4}$ hidden nodes in the first and second hidden layers. A configuration with three hidden layers, of sizes $128, 32, 8$, was also tested, for which $X$, $Y$ were the activation vectors from layers $1$ and $3$. 

\paragraph*{\em Implementation}
%(wdbc, optdigits, splice, wine, tic-tac-toe, dna) and regression 
%(boston, Diabetes(scikit-learn), stock, balloon, cpu, cloud). 
NN were trained using the \verb|MLPClassifier| and \verb|MLPRegressor| classes in \verb|scikit-learn|~\cite{scikit-learn}, with cross-entropy or quadratic loss for classification and regression, respectively, ReLU activation functions, Glorot-He pseudorandom initialization~\cite{GlorotBengio2010},~\cite{HeEtAl2015}, Adam optimizer~\cite{Adam2015}, learning rate of $0.001$, and a maximum of $2000$ training iterations. CKA was implemented in Python (\url{https://docs.python.org/3/license.html}), using NumPy~\cite{NumPy2020} (\url{https://numpy.org/doc/stable/license.html}) and Matplotlib~\cite{Matplotlib2007} (PSF license, \url{https://docs.python.org/3/license.html}). Experiments were performed on a workstation with an Intel i9-7920X (12 core) processor and 128GB RAM, under Ubuntu 18.04.5 LTS (GNU public license). 

\paragraph*{\em Experiments}
$50$ runs were performed for each pair $(D,w)$ of a data set $D$ and NN width $w$. In each run, a new NN model was trained on the full data set, starting from fresh pseudorandom initial parameter values. A training run was repeated if the resulting in-sample accuracy was below $0.8$ (classification) or if in-sample coefficient of determination was below $0.5$ (regression), but not if the optimization had not converged within the allowed number of iterations. After the network had been trained in a given run, $\text{CKA}_{\text{lin}}$ was computed once, and $\text{CKA}_{G(\sigma)}$ was computed for each bandwidth $\sigma = 2^p,\; p = -4, -3, \cdots, 8$, for a total of $50$ $\text{CKA}_{\text{lin}}$ and $650$ $\text{CKA}_{G(\sigma)}$ evaluations per $(D,w)$ pair. 

\paragraph*{\em Compute time}
Compute time for the results presented in the paper was approximately $32$ hours, 
much of it on the $8 \cdot 12 \cdot (50 + 650) = 67200$ CKA evaluations needed across the $8$ network
widths (including the 128-32-8 three-hidden-layer configuration) and the $12$ data sets. 
Additional runs for validation required another $12$ hours. Several shorter preliminary runs were carried out for debugging and initial selection of the NN hyperparameters; configurations and threshold values were selected empirically in order to ensure a successful end to training after no more than a handful of runs in nearly all cases. Total compute time across all runs is estimated to have been $50$ hours.

\paragraph*{\em Evaluation metrics}
CKA means and standard errors (SE = standard deviation divided by $\sqrt{50}$), and medians and standard error equivalent (SE = inter-quartile range divided by $\sqrt{50}$) of the ratio $\rho$ of Eq.~\ref{eq:rho} were computed across runs. We measured the magnitude of the discrepancy between $\text{CKA}_{G(\sigma)}$ and $\text{CKA}_{\text{lin}}$ by the base-$2$ logarithm of the relative difference between the two measures as in
Eq.~\ref{eq:logRelDifference}.
\begin{equation}
\log_2 \text{rel. CKA difference} 
= \log_2 \left ( \frac{|\text{CKA}_{G(\sigma)} - \text{CKA}_{\text{lin}}|}{\text{CKA}_{\text{lin}}} \right )
\label{eq:logRelDifference}
\end{equation}
Theorem~\ref{thm:limCKAGequalsCKAL} implies that, for large $\sigma$, the logarithmic relative difference
of Eq.~\ref{eq:logRelDifference} should decrease along a straight line of slope $-2$ as a function of $\log_2 \sigma$, reflecting a $O(1/\sigma^2)$ dependence. Accordingly, for each data set, we determined a $1/\sigma^2$ asymptote by extrapolating backward from the largest tested bandwidth value,
$\sigma = 2^8$. If $\sigma < 2^8$, the predicted relative difference along the $1/\sigma^2$ asymptote is as in Eq.~\ref{eq:relDiffAsymptote}, where $r_8$ is the observed relative difference at $\sigma = 2^8$.
\begin{equation}
\text{predicted } \log_2 (\text{rel. CKA diff. at } \sigma) = \log_2 r_8 - 2(\log_2 \sigma - 8)
\label{eq:relDiffAsymptote}
\end{equation}
We determined a $1/\sigma^2$ convergence onset bandwidth, $\sigma^*_0$, as the minimum bandwidth above which the observed $\log_2$ relative difference between Gaussian and linear CKA differs by less than $0.25$ from the predicted value along that data set's $1/\sigma^2$ asymptote (Eq.~\ref{eq:convergenceOnsetBandwidth}).
\begin{equation}
\sigma^*_0 = \min \{\sigma_0 \mid \text{log rel.\ diff.}(\sigma) < 0.25 \text{ for all } \sigma \ge \sigma_0 \}
\label{eq:convergenceOnsetBandwidth}
\end{equation}
%The threshold value $0.25$ in Eq.~\ref{eq:convergenceOnsetBandwidth} ensures a difference less than $20\%$.
Threshold values other than $0.25$ in Eq.~\ref{eq:convergenceOnsetBandwidth} ($1, 0.5, 0.1$), corresponding
to different tolerances for the log relative difference, yielded similar results in most cases in terms of the resulting relative $\sigma^*_0$ ranks of the different data sets.

\subsection{Discussion of experimental results} % (additional results tables in the Appendix)}
\label{section:expResults}

\begin{table*}[h!]
\caption{Median ratios, $\rho = \max(\text{diam}(X)/d_X, \text{diam}(Y)/d_Y)$, and $\pm 2$ standard error equivalent confidence intervals,\\of maximum to median distance between features. $w$ is NN width. Regression.}
% 50 reps
\centering
\begin{tabular}{ccccccc}
\toprule
$w$ &cpu &boston &Diab(skl) &stock &balloon &cloud\\
\midrule
%$\rho \pm 2\frac{\text{IQR}}{\sqrt{50}}$ 
16 &10.63 $\pm$ 0.53 &5.08 $\pm$ 0.54 &4.87 $\pm$ 0.01 &3.09 $\pm$ 0.28 &14.62 $\pm$ 0.88 &6.59 $\pm$ 0.45\\
32 &9.60  $\pm$ 0.58 & 3.87 $\pm$ 0.23 & 4.86 $\pm$ 0.01 & 2.68 $\pm$ 0.14 &15.64 $\pm$ 0.83 & 5.76 $\pm$ 0.38\\
64 &9.23 $\pm$ 0.29 & 3.57 $\pm$ 0.29 & 4.86 $\pm$ 0.01 & 2.62 $\pm$ 0.12 &16.25 $\pm$ 0.90 & 5.99 $\pm$ 0.26\\
128& 9.15 $\pm$ 0.20 & 3.61 $\pm$ 0.11 & 4.85 $\pm$ 0.01 & 2.50  $\pm$ 0.05 &18.49 $\pm$ 0.86 & 5.84 $\pm$ 0.19\\
256 &9.19 $\pm$ 0.16 & 3.40  $\pm$ 0.07 & 4.84 $\pm$ 0.01 & 2.48 $\pm$ 0.05 &20.70  $\pm$ 0.88 & 6.05 $\pm$ 0.10\\
%0.26, 0.27, 0.  , 0.14, 0.44, 0.22	% twice the third value is 0.01 to two digits
%0.53, 0.54, 0.01, 0.28, 0.88, 0.45  % two IQR/sqrt(50) on this row
512 &8.91 $\pm$ 0.12 &3.38 $\pm$ 0.04 &4.84 $\pm$ 0.01 &2.44 $\pm$ 0.04 &22.65 $\pm$ 0.72 &6.20 $\pm$ 0.12\\
1024 &8.82 $\pm$ 0.11 &3.42 $\pm$ 0.03 &4.82 $\pm$ 0.04 &2.43 $\pm$ 0.02 &25.18 $\pm$ 1.14 &6.16 $\pm$ 0.19\\
\midrule
128-32-8	&9.96 $\pm$ 0.54 &3.91 $\pm$ 0.24 &4.83 $\pm$ 0.01 &3.11 $\pm$ 0.25 &22.71 $\pm$ 2.97 &7.30 $\pm$ 0.23\\
\bottomrule
\end{tabular}
\normalsize
\label{table:regrRhoMeansIQR}
\end{table*}

The results show a range of geometric characteristics of learned feature
representations across data sets, seen as differences in the representation eccentricity, $\rho$,
defined in Eq.~\ref{eq:rho}.
%, of maximum to median distance between feature vectors. 
Geometry is stable for a given data set, as indicated
by a small standard deviation for $\rho$, with some dependence on network size. See Table~\ref{table:regrRhoMeansIQR} for the case of regression; the Appendix includes
the information for classification.

%\paragraph*{\em Representation dimensionality influences the eccentricity ratio, $\rho$}
%, of maximum to median feature distance}
We observe generally lower values of $\rho$ (Eq.~\ref{eq:rho})
for wider networks in Table~\ref{table:regrRhoMeansIQR}.
%(additional information in the Appendix). 
%(Tables~\ref{table:classRhoMeansIQR} and~\ref{table:regrRhoMeansIQR}). 
This is consistent with expectations, as concentration of Euclidean distance in high dimensions~\cite{AggarwalEtAlHighDDistance2001}
will bring the ratio of maximum to median distance between feature vectors
closer to $1$ as network width grows. 
%as maximum and median distance between features will be closer in a higher-dimensional space. 
The balloon data set is the only one for which $\rho$ increases
with network width. That data set contains a small group of examples with lower values of the
two attributes than the rest (indices 332,  706,  969, 1025, 1041, 1399, 1453, 1510). Those examples are increasingly distant from the majority (in units of median distance) in the deeper (second or third) hidden layer representations as width increases, driving the increase in $\rho$;
the ratio of maximum to median distance in the first hidden layer representation
does not exhibit similar growth.

Dimensionality of the raw data sets (Table~\ref{table:datasetSizes}) appears to filter into
the learned neural representations, % for data sets other than balloon, 
as well. For a fixed network width, the correlation between raw data dimensionality and $\rho$ is unambiguously negative, between $-0.38$ and $-0.33$ for classification, and between $-0.8$ and $-0.85$ for regression, even though
representation dimensionality is fixed by network width.

\paragraph*{\em Noticeable differences between Gaussian and linear CKA occur for small bandwidths}
Mean relative difference values between $\text{CKA}_{G(\sigma)}$ and $\text{CKA}_{\text{lin}}$ greater than $0.2$ ($\log_2$ values greater than $-2.3$), are observed
for several data sets when $\sigma \le 0.25$ (when $\log_2 \sigma \le -2$) in the case of classification; in fact, relative CKA difference values greater than $0.7$ ($\log_2 \text{ rel.\ CKA diff.} > -0.5$, values in the Appendix) are observed for the dna data set when network width $w$ is $128$ or greater. This shows that noticeably nonlinear behavior of Gaussian CKA is quite possible for small bandwidth values. 
For some classification data sets, however (wdbc, wine), and most regression data sets, the relative CKA difference remains comparatively small for all bandwidths.

\comment{%moved to Future Work
In light of this evidence, general assessments of CKA as a representational similarity metric based on linear CKA alone, under the assumption that Gaussian RBF CKA would yield equivalent results (e.g.,~\cite{DingEtAlNeurIPS2021}) may benefit from a closer quantitative examination. 
%regardless of feature representation characteristics. 
}%moved to Future Work

\begin{figure}[h]
	\begin{center}
		\includegraphics [width=0.51\columnwidth, clip=true, trim=2mm 0mm 14mm 14mm]
		{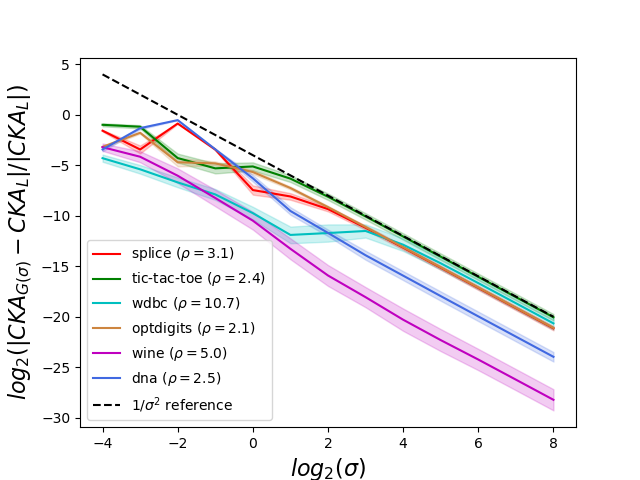}
		\includegraphics [width=0.475\columnwidth, clip=true, trim=12mm 0mm 14mm 14mm]
		{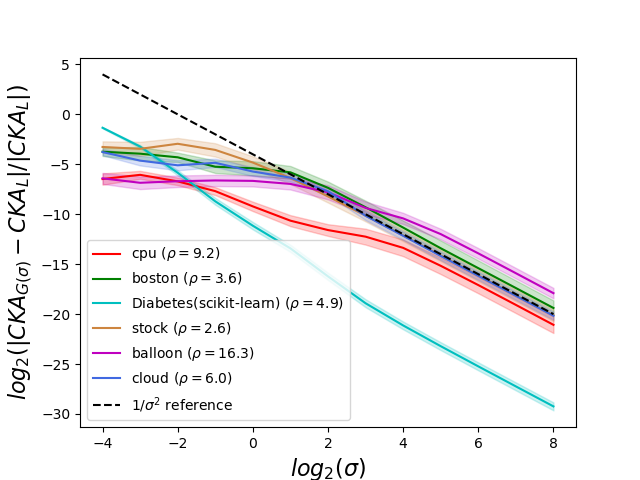}
	\end{center}
	\caption{Relative difference between Gaussian and linear CKA for neural feature representations of classification (left) and regression (right) data sets. Neural network width is $w=64$. Shading extends two standard errors from the mean. Dotted reference line of slope $-2$ indicates $1/\sigma^2$ relationship. Gaussian CKA (bandwidth $\sigma$) converges to linear CKA like $1/\sigma^2$ as $\sigma \rightarrow \infty$, for all network widths. Onset of $1/\sigma^2$ convergence is delayed for representations of high eccentricity, $\rho$ (Table~\ref{table:sigma0Stars}).}
	\label{fig:CKAGaussianLinearDifferenceEmpirical}
\end{figure}

\begin{figure}[h]
	\begin{center}
		\includegraphics [width=0.51\columnwidth, clip=true, trim=2mm 0mm 14mm 14mm]
		{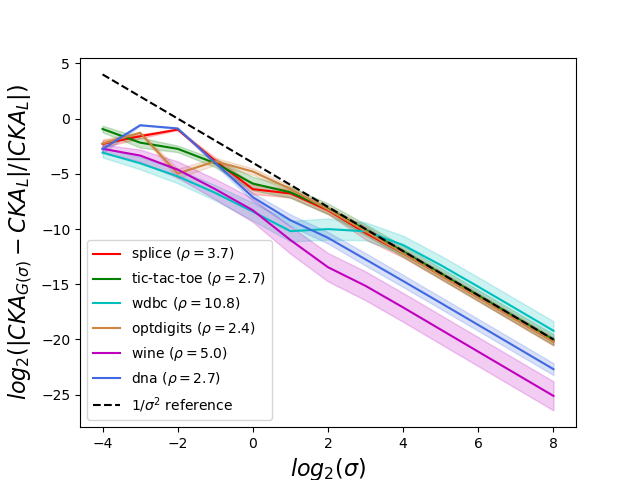}
		\includegraphics [width=0.475\columnwidth, clip=true, trim=12mm 0mm 14mm 14mm]
		{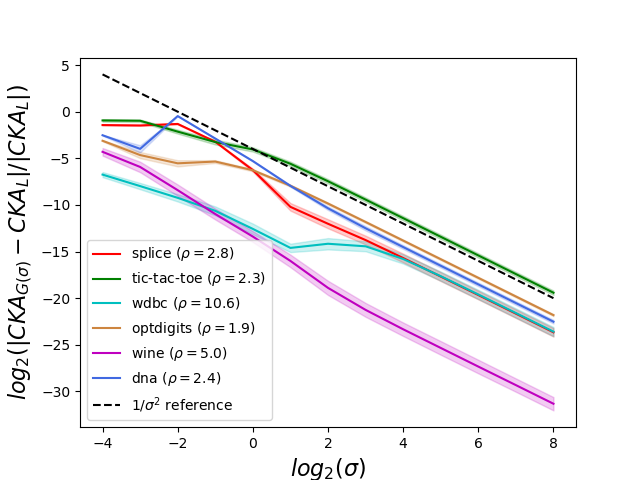}
	\end{center}
	\caption{Relative difference between Gaussian and linear CKA for neural feature representations associated with networks of different widths $w=32$ (left) and $w = 256$ (right). Classification data sets. Shading extends two standard errors from the mean. $1/\sigma^2$ 
	convergence is observed in both cases, as well as for all tested widths not shown. Wider networks are less sensitive to initial parameter values, as evidenced by lower standard error of the CKA relative difference;
	values appear in the Appendix.}
	\label{fig:CKAGaussianLinearDifferenceWiderNetworks}
\end{figure}

\paragraph*{\em Gaussian CKA converges to linear CKA like $1/\sigma^2$}
%Table~\ref{table:dataSetDiameters} shows the means and standard errors of the data set diameters.
Fig.~\ref{fig:CKAGaussianLinearDifferenceEmpirical} shows confidence intervals for the means, of radius two standard errors, of the observed $\log_2$ relative difference between $\text{CKA}_{G(\sigma)}$ and $\text{CKA}_{\text{lin}}$ as a function of Gaussian bandwidth, $\sigma$ (Eq.~\ref{eq:logRelDifference}). 
%Numerical values in Supplementary Tables~\ref{table:classRelDiffMeans},~\ref{table:regrRelDiffMeans}.
Convergence at the rate $1/\sigma^2$ for $\sigma \gg 1$ is observed
(log-log slope of $-2$ in Fig.~\ref{fig:CKAGaussianLinearDifferenceEmpirical}), 
as described in Theorem~\ref{thm:limCKAGequalsCKAL}. Results are stable across
the range of neural network configurations tested. Standard error of the relative CKA difference is observed to decrease as network width increases (e.g., Fig.~\ref{fig:CKAGaussianLinearDifferenceWiderNetworks}), suggesting that wider networks are less sensitive to variations in initial parameter values.

The Diabetes(scikit-learn) data set stands out for having, by far, the smallest mean relative CKA difference among data sets tested, across much of the $\sigma$ range. 
The neural feature maps $X$ and $Y$ for that data set have nearly proportional linear Gram matrices $K_{\text{lin}}$ and $L_{\text{lin}}$, hence the linear CKA value is very close to $1$. Because the matrix of squared inter-example distances depends linearly on the linear Gram matrix, the Gaussian CKA value is also very close to $1$.

\paragraph*{\em The representation eccentricity $\rho$ (Eq.~\ref{eq:rho}) is reflected in convergence onset}
Our experimental results suggest that noticeably
nonlinear behavior of Gaussian CKA occurs almost exclusively for bandwidths $\sigma < \rho$.
Indeed, for all data sets tested, the observed mean relative difference between linear and Gaussian CKA
is less than $0.01$ when $\sigma > \rho$, where $\rho$ is the ratio of maximum to median distance between feature vectors (Eq.~\ref{eq:rho}). 
%See Table~\ref{table:classRhoMeansIQR} and Table~\ref{table:regrRhoMeansIQR}.
This  %in Fig.~\ref{fig:CKAGaussianLinearDifferenceEmpirical} 
confirms our finding in section~\ref{subsection:dataDependentSigmaBound} that 
$\text{CKA}_G(\sigma)$ differs little from 
$\text{CKA}_{\text{lin}}$ if $\sigma \gg \rho$. 
Furthermore, our results suggest that the threshold between nonlinear and linear regimes of Gaussian
CKA is not substantially less than $\rho$: while mean relative CKA difference for $\sigma \ge 1$ peaked under $0.1$ for two-hidden-layer neural representations, across all data sets tested, relative difference values less than $0.1$ occur for the tic-tac-toe data set in the three-layer 128-32-8 representation only when $\sigma \ge 4$; the lower end of
this bandwidth range is quite close to the median $\rho$ value of $4.2$ for that representation (which differs from the width-$64$ two-layer representation in Figs.~\ref{fig:CKAGaussianLinearDifferenceEmpirical},~\ref{fig:CKAGaussianLinearDifferenceWiderNetworks}).

We find additional support for the view that $\rho$ approximates the boundary between nonlinear and linear regimes of Gaussian CKA in the fact that the convergence onset bandwidth, $\sigma^*_0$ (Eq.~\ref{eq:convergenceOnsetBandwidth}),
%for each data set to be the minimum bandwidth value such that the vertical distance in Fig.~\ref{fig:CKAGaussianLinearDifferenceEmpirical} between the relative CKA difference plot and the $1/\sigma^2$ asymptote for that data set is less than $0.25$ whenever $\sigma \ge \sigma^*_0$, we find that $\sigma^*_0$ 
is largest for the data sets of largest $\rho$: median $\log_2 \sigma^*_0 = 5$, $4$, $4$ for balloon ($\rho \approx 16$), wdbc ($\rho \approx 11$), cpu ($\rho \approx 9$), respectively ($\sigma^*_0$ and $\rho$ values differ slightly among network widths and depths, but data sets of largest $\rho$ are the same). See Table~\ref{table:sigma0Stars}.
Diabetes(scikit-learn) ties for third with $\log_2 \sigma^*_0 = 4$; median $\log_2 \sigma^*_0 < 4$ for all other data sets.
%(cloud is the most sensitive among data sets tested to the choice of threshold value in the $\sigma^*_0$ definition: its $\sigma^*_0$  drops from $4$ to $0$ if a threshold of $0.5$ is used instead of $0.25$; $\sigma^*_0$ changes by at most $1$ for other data sets). 
Large differences in $\rho$ between data sets generally translate directly to differences between their $\sigma^*_0$ values. This is consistent with the analysis in section~\ref{subsection:dataDependentSigmaBound}, as $\rho$ also controls terms of orders
higher than $2$ in the exponential power series.

Our experimental results as a whole confirm $1/\sigma^2$ convergence of $\text{CKA}_{G(\sigma)}$ to $\text{CKA}_{\text{lin}}$ as $\sigma \rightarrow \infty$ (Theorem~\ref{thm:limCKAGequalsCKAL}), and support the use of the representation eccentricity, $\rho$ (Eq.~\ref{eq:rho}), as a useful heuristic upper bound on the range of bandwidths for which Gaussian CKA can display nonlinear behavior.

\begin{table*}[h!]
\caption{Log bandwidth, $\log_2 \sigma^*_0$, of $1/\sigma^2$ convergence onset. 
$w$ denotes network width.}	% 50 reps
\centering
\begin{tabular}{cccccccclcccccc}
\toprule
&\multicolumn{6}{c}{Classification} &&\multicolumn{6}{c}{Regression}\\
$w$ &splice &t-t-t &wdbc &optd &wine &dna &&cpu &bost &diab &stock &balln &cloud\\
\cmidrule{2-7} \cmidrule(l){9-14}
16 &2 &2 &5 &2 &-1 &2 &&4 &3 &4 &1 &5 &4\\	% threshold 0.25
32 &2 &2 &4 &1 &3 &2 &&4 &3 &4 &1 &5 &4\\
64 &2 &2 &4 &1 &3 &3 &&5 &2 &4 &2 &5 &3\\
128 &3 &1 &4 &1 &2 &2 &&4 &2 &4 &1 &5 &2\\
256 &3 &1 &4 &1 &3 &2 &&5 &2 &4 &1 &5 &3\\
512 &3 &1 &4 &1 &3 &2 &&4 &2 &4 &1 &5 &2\\
1024 &2 &1 &4 &1 &3 &2 &&5 &2 &5 &1 &5 &2\\
\midrule
%\cmidrule{2-7} \cmidrule(l){9-14}
128-32-8 &3 &2 &4 &1 &2 &2	&&4 &2 &3 &0 &6 &3\\
\bottomrule
\end{tabular}
\normalsize
\label{table:sigma0Stars}
\end{table*}

\section{Conclusions}

This paper considered the large-bandwidth asymptotics of CKA using Gaussian RBF kernels.
We proved rigorously that mean-centering of the feature vectors 
ensures that Gaussian RBF CKA converges to linear CKA in the large bandwidth limit, with an $O(1/\sigma^2)$ asymptotic relative difference between them. We showed that a similar result does not hold for the non-centered kernel alignment measure of~\cite{CristianiniEtAl2001}. 

We also showed that the geometry of the feature representations impacts the range of bandwidths for
which Gaussian CKA can behave nonlinearly, focusing on the representation eccentricity ratio, $\rho$, of maximum to median distance between feature vectors as a representation-sensitive guide. 
Our results suggest that bandwidth values less than $\rho$ can lead to noticeably nonlinear behavior of Gaussian CKA, whereas bandwidths larger than $\rho$ will yield essentially linear behavior. In order to enable nonlinear modeling, the bandwidth should, therefore, be selected in the interval $(0,\rho)$. 
%larger bandwidths will generally produce equivalent results to linear CKA.

\section{Future Work}
\label{section:futureWork}

Representation eccentricity, $\rho$, %described in this paper 
correlates well with the bandwidth at which Gaussian CKA transitions between nonlinear and linear regimes,
and our theoretical result establishes convergence of Gaussian to linear CKA for large bandwidths.
In applications, however, the order of magnitude of the difference between Gaussian and linear CKA can be
of greatest interest in regard to kernel selection. %is small even for small bandwidths. 
One direction for future work would be to seek additional representation characteristics that, in conjunction with eccentricity, can better gauge the order of magnitude of the Gaussian-linear CKA difference for a given representation. Such work could provide further guidance in selecting between Gaussian
and linear CKA kernels for particular %representation types and 
applications.
Robust versions of the representation eccentricity can also be explored, in which maximum and median
distance are replaced by other pairs of quantiles of the distance distribution.

%a more complete picture of the nonlinear behavior of Gaussian CKA at a given bandwidth.
%, and which could further aid in selecting between Gaussian and linear kernels for given feature %representations. 
% may also be beneficial to pursue a closer quantitative examination of
%general assessments of CKA as a similarity metric based on linear CKA alone (e.g.~\cite{DingEtAlNeurIPS2021}),
%given that the results of the present paper suggest that noticeable differences between Gaussian CKA
%and linear CKA can occur for bandwidths much smaller than $\rho$.

\section*{Declarations}
The work for this paper did not involve human subjects. Data sets used do not include any
personally identifiable information or offensive content. The author has no conflicts of interest to report.

\comment{ % Broader Impact
\section*{Broader impact}
As with other machine learning techniques, kernel similarity poses a potential risk if applied blindly. Similarity values should not be interpreted as certifying the degree to which two
alternative data representation approaches are equivalent for all purposes; one of them may require substantially greater computation, for example, with greater negative environmental effects. 
Likewise, similarity results are entirely dependent on the set of data attributes selected initially as descriptors of individual examples. Bias or incompleteness of this initial description, or in subsequent representation learning, can persist in the resulting similarity assessments. These considerations are especially important for data that involve information about human individuals or groups. No data representation or computation should be viewed as providing a legitimate basis for establishing human value.
} % Broader Impact

\ifCLASSOPTIONcompsoc
  \section*{Acknowledgments}
\else
  \section*{Acknowledgment}
\fi

The author thanks Sarun Paisarnsrisomsuk, whose experiments for~\cite{SarunPhDThesis2021}
motivated this paper, and Carolina Ruiz, for helpful comments on an earlier version of
the manuscript.

%\newpage

\small
\bibliography{IEEEabrv, CKAReferences}
\normalsize

%%%%%%%%%%%%%%%%%%%%%%%%%%%%%%%%%%%%%%%%%%%%%%%%%%%%%%%%%%%%

%\comment{ %Appendix / supplementary material

%\newpage
%{\ }
%\newpage

\bigskip

\appendices
\section{Supplementary information for the paper\\
{\normalfont Gaussian RBF CKA 
in the large bandwidth limit}}

\subsection{Details of the counterexample for non-centered kernel alignment in section~\ref{subsection:GaussianCKAConvergesToLinearCKA}}

As we state in section~\ref{subsection:GaussianCKAConvergesToLinearCKA}, Theorem~\ref{thm:limCKAGequalsCKAL} hinges on the assumption that the Gram matrices are mean-centered. We justify that statement here. We show that an analogous convergence result does not hold for the
non-centered kernel alignment of~\cite{CristianiniEtAl2001}, by considering the feature matrices below.
\begin{align*}
X &= \begin{bmatrix}
1 &0\\
0 &1
\end{bmatrix}
\ 
&Y = \begin{bmatrix}
1 &0\\
1 &1
\end{bmatrix}
\end{align*}
In order to compute $\text{CKA}(K_{G(\sigma)},L_{G(\sigma)})$, we first compute, for each of the feature matrices $X$ and $Y$, the corresponding matrix of distances between pairs of feature vectors (rows):
\begin{align*}
D(X) &= \begin{bmatrix}
0 &\sqrt{2}\\
\sqrt{2} &0
\end{bmatrix}
\ 
&D(Y) = \begin{bmatrix}
0 &1\\
1 &0
\end{bmatrix}
\end{align*}
$D(X)$ and $D(Y)$ are scalings of one another. 
Given a bandwidth, $\sigma$, we
interpret $\sigma$ as expressed in units of median distance, as in the note in section~\ref{subsection:kernelsAndGramMatrices}. If diagonal zeros are included
in the median computation, then
$\sigma_X = d_X \sigma = \frac{\sigma}{\sqrt{2}}$ in the non-centered Gaussian entries $e^{-\frac{|x_i-x_j|^2}{2\sigma^2_X}}$ of $K_{G(\sigma)}$, and $\sigma_Y = d_Y \sigma = \frac{\sigma}{2}$ in the non-centered Gaussian entries $e^{-\frac{|y_i-y_j|^2}{2\sigma^2_Y}}$ of $L_{G(\sigma)}$. Median scaling makes the non-centered matrices $K_{G(\sigma)}(X)$ and $L_{G(\sigma)}(Y)$ identical:
\begin{align*}
K_{G(\sigma)}(X) &= 
\begin{bmatrix}
1 &e^{-\frac{2}{\sigma^2}}\\
e^{-\frac{2}{\sigma^2}} &1
\end{bmatrix}
= L_{G(\sigma)}(Y)
\end{align*}
Thus, numerator and denominator of the non-centered CKA expression are also equal,
so Gaussian CKA has the value $1$:
\begin{align}
\text{CKA}(K_{G(\sigma)}, L_{G(\sigma)})
&=
\frac{tr \left ( K_{G(\sigma)} L_{G(\sigma)}\right )}
{\sqrt{tr \left ( K_{G(\sigma)} K_{G(\sigma)}\right ) tr \left ( L_{G(\sigma)} L_{G(\sigma)}\right )}} \nonumber
\\
&= \frac{ \left (1 + e^{-\frac{2}{\sigma^2}} \right )^2 }
{\sqrt{\left (1 + e^{-\frac{2}{\sigma^2}} \right )^2 \left (1 + e^{-\frac{2}{\sigma^2}} \right )^2}}
=
1
\label{eq:ceGaussCKA}
\end{align}
Different intermediate numerical values occur if diagonal zeros in $D(X)$ and $D(Y)$ are excluded from the median computation, % ($\sigma_X = d_X \sigma = \sqrt{2}\sigma$, $\sigma_Y = d_Y \sigma = \sigma$), 
but the final result in Eq.~\ref{eq:ceGaussCKA} is the same.

Now consider a linear kernel. First, compute the Gram matrices of pairwise dot products between rows:
\begin{align*}
K_{\text{lin}}(X) &=
\begin{bmatrix}
1 &0\\
0 &1
\end{bmatrix}
\ 
&L_{\text{lin}}(Y) =
\begin{bmatrix}
1 &1\\
1 &2
\end{bmatrix}
\end{align*}
We find non-centered linear CKA straight from the definition:
\begin{align}
\text{CKA}(K_{\text{lin}}, L_{\text{lin}})
&=
\frac{tr \left ( K_{\text{lin}} L_{\text{lin}}\right )}
{\sqrt{tr \left ( K_{\text{lin}} K_{\text{lin}}\right ) tr \left ( L_{\text{lin}} L_{\text{lin}}\right )}} \nonumber
\\
&= \frac{ 1 + 2 }
{\sqrt{\left ( 1 + 1 \right ) \left ( 2 + 5 \right )}}
=
\frac{3}{\sqrt{14}}
\label{eq:ceLinCKA}
\end{align}
Eqs.~\ref{eq:ceGaussCKA} and~\ref{eq:ceLinCKA} show that, without centering, Gaussian and linear CKA
remain at a fixed positive distance from one another in this example for all bandwidths $\sigma$.
This proves that an analog of Theorem~\ref{thm:limCKAGequalsCKAL} fails for
non-centered CKA.
%\boxenddemo

%\newpage

\subsection{Proof of Theorem~\ref{thm:limCKAGequalsCKAL} in case of two Gaussian kernels}

The proofs in the paper focus on the case in which one kernel is Gaussian with bandwidth $\sigma \rightarrow \infty$ and the other is a fixed kernel, $L$. We show here how, as indicated at the end of the proof of Lemma~\ref{lemma:limCKAGequalsCKAE}, the case of two Gaussian kernels with large bandwidths follows by a triangle inequality argument. We will abbreviate $\text{CKA}(K,L)$ as
$\text{C}(K,L)$; note that the feature representations in $K$, $L$ are $X$, $Y$, respectively.

First, we address Lemma~\ref{lemma:limCKAGequalsCKAE} in the case of two Gaussian kernels. Begin by decomposing the target difference between Gaussian and Euclidean CKA as follows:
\small
\begin{align*}
&\text{C}(K_{G(\sigma_1)}, L_{G(\sigma_2)}) - \text{C}(K_E, L_E) =\\ 
&\text{C}(K_{G(\sigma_1)}, L_{G(\sigma_2)}) - \text{C}(K_E, L_{G(\sigma_2)})
+ \text{C}(K_E, L_{G(\sigma_2)}) - \text{C}(K_E, L_E)
\end{align*}
\normalsize

Given any desired tolerance, $\epsilon > 0$, the single-Gaussian case of Lemma~\ref{lemma:limCKAGequalsCKAE} as proved in the main text 
(together with symmetry of CKA in its two arguments) 
shows that there exists a bandwidth $\underline{\sigma_2}$, such that the difference term at 
far right, above, is less than $\epsilon/2$ in absolute value whenever 
$\sigma_2 > \underline{\sigma_2}$.

Having fixed the bandwidth $\underline{\sigma_2}$, there similarly exists a 
bandwidth $\underline{\sigma_1}$ such that the absolute value of the first difference term 
on the right-hand side above is less than $\epsilon/2$ whenever
$\sigma_1 > \underline{\sigma_1}$. By the scalar triangle inequality, it now
follows that the absolute value of the target difference on the left-hand side
above is smaller than $\epsilon$ whenever both $\sigma_1 > \underline{\sigma_1}$ and 
$\sigma_2 > \underline{\sigma_2}$.

This argument proves Lemma~\ref{lemma:limCKAGequalsCKAE} in the case of two Gaussian kernels:
as $\sigma_1,\ \sigma_2 \rightarrow \infty$,
\begin{align}
\text{C}(K_{G(\sigma_1)}, L_{G(\sigma_2)}) &= 
\text{C}(K_E, L_E)\ +\ O \left ( \frac{1}{\min(\sigma_1, \sigma_2)^2} \right )
%\qquad \text{as} \quad\sigma_1,\ \sigma_2 \rightarrow \infty
\label{eq:Lemma1TwoGaussiansVersion}
\end{align}

The Corollary to Lemma~\ref{lemma:HSICEequalsNegative2HSICL} holds for any kernel $L$, including a
Euclidean kernel, $L_E$:
\begin{align*}
\text{C}(K_E, L_E) &= \text{C}(K_{\text{lin}}, L_{\text{lin}})
\end{align*}
Therefore, the case of two Gaussian kernels in Theorem~\ref{thm:limCKAGequalsCKAL} now follows
from Eq.~\ref{eq:Lemma1TwoGaussiansVersion}:
\begin{align*}
\text{C}(K_{G(\sigma_1)}, L_{G(\sigma_2)}) &= 
\text{C}(K_{\text{lin}}, L_{\text{lin}})\ +\ O \left ( \frac{1}{\min(\sigma_1, \sigma_2)^2} \right )
%\qquad \text{as} \quad\sigma_1,\ \sigma_2 \rightarrow \infty
\end{align*}
%\boxenddemo

\subsection{CKA difference for three-layer NN architecture}

\begin{figure}[h]
	\begin{center}
		\includegraphics [width=0.51\columnwidth, clip=true, trim=2mm 0mm 14mm 14mm]
		{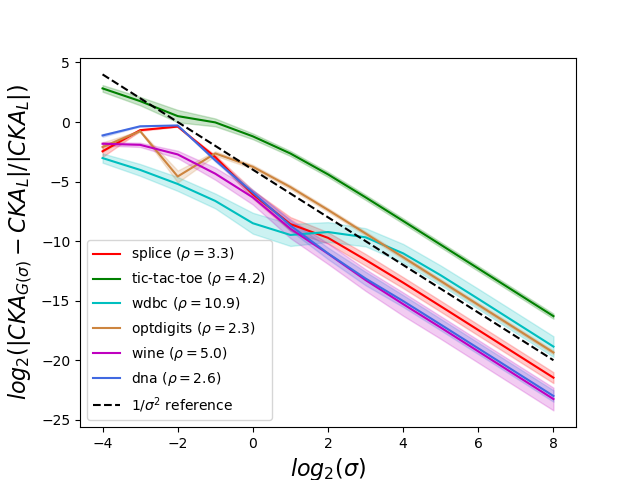}
		\includegraphics [width=0.475\columnwidth, clip=true, trim=12mm 0mm 14mm 14mm]
		{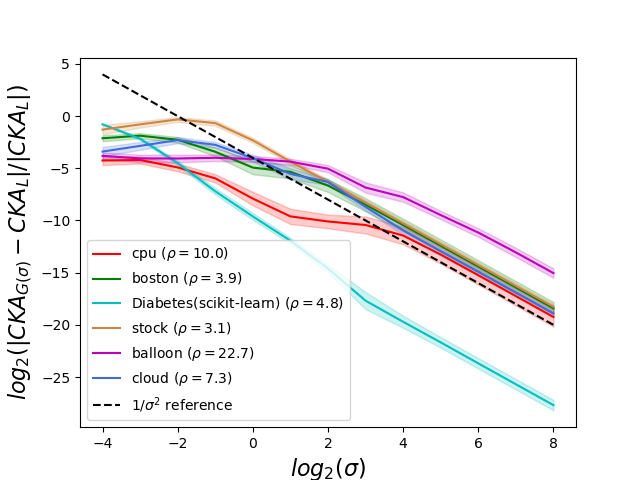}
	\end{center}
	\caption{Relative difference between Gaussian and linear CKA for neural feature representations of classification (left) and regression (right) data sets. Three-layer 128-32-8 neural network configuration. Shading extends two standard errors from the mean. Dotted reference line of slope $-2$ indicates $1/\sigma^2$ relationship. Gaussian CKA (bandwidth $\sigma$) is observed to converge to linear CKA like $1/\sigma^2$ as $\sigma \rightarrow \infty$. Onset of $1/\sigma^2$ convergence is delayed for data sets of large $\rho$ (see text).}
	\label{fig:CKAGaussianLinearDifferenceThreeHiddenLayers}
\end{figure}

%\newpage

\subsection{Results tables for section~\ref{section:expResults} experiments}

\begin{table}[h!]
\caption{Mean rel.\ difference $\log_2(|\text{CKA}_{G(\sigma)}-\text{CKA}_{\text{lin}}|/\text{CKA}_{\text{lin}})$. \\NN width $w=16$. Classification.}	% 50 reps
\centering
\begin{tabular}{ccccccc}
\toprule
$\log_2 \sigma$ &splice &tic-tac-toe &wdbc &optdigits &wine &dna\\
\midrule
-4 &-1.28 &-1.34 &-2.25 &-0.83 &-2.21 &-0.80\\
-3 &-0.54 &-2.04 &-3.03 &-1.75 &-2.54 &-0.40\\
-2 &-1.29 &-2.12 &-4.08 &-3.66 &-3.58 &-1.51\\
-1 &-3.99 &-3.12 &-5.46 &-2.92 &-4.85 &-4.64\\
0 &-5.52 &-4.29 &-6.70  &-3.70  &-6.87 &-6.21\\
1 &-5.78 &-5.41 &-8.33 &-4.97 &-8.81 &-6.84\\
2 &-7.30  &-6.88 &-8.12 &-6.70 &-11.00 &-8.50\\
3 &-9.19 &-8.70  &-8.96 &-8.63 &-12.83 &-10.42\\
4 &-11.17 &-10.66 &-9.97 &-10.61 &-14.82 &-12.39\\
5 &-13.16 &-12.64 &-11.70  &-12.61 &-16.83 &-14.39\\
6 &-15.16 &-14.64 &-13.64 &-14.61 &-18.84 &-16.39\\
7 &-17.16 &-16.64 &-15.63 &-16.61 &-20.84 &-18.39\\
8 &-19.16 &-18.64 &-17.62 &-18.61 &-22.84 &-20.39\\
\bottomrule
\end{tabular}
\normalsize
\label{table:classRelDiffMeans}
\end{table}

\begin{table}[h!]
\caption{Mean rel.\ difference $\log_2(|\text{CKA}_{G(\sigma)}-\text{CKA}_{\text{lin}}|/\text{CKA}_{\text{lin}})$. \\NN width $w=32$. Classification.}	% 50 reps
\centering
\begin{tabular}{ccccccc}
\toprule
$\log_2 \sigma$ &splice &tic-tac-toe &wdbc &optdigits &wine &dna\\
\midrule
-4  &  -2.29&  -0.94&  -3.08&  -2.27&  -2.73&  -2.77\\
-3  &  -1.60&  -2.18&  -4.05&  -1.28&  -3.35&  -0.60\\
-2  &  -0.99&  -2.74&  -5.23&  -4.98&  -4.62&  -0.89\\
-1  &  -3.83&  -4.04&  -6.72&  -3.87&  -6.39&  -4.01\\
0  &  -6.39&  -5.88&  -8.43&  -4.80&  -8.30&  -7.12\\
1  &  -6.78&  -6.68& -10.19&  -6.37& -10.99&  -9.20\\
2  &  -8.30&  -8.17& -10.01&  -8.24& -13.46& -10.80\\
3  & -10.35& -10.05& -10.20& -10.20& -15.15& -12.75\\
4  & -12.19& -12.03& -11.45& -12.19& -17.1 & -14.71\\
5  & -14.18& -14.02& -13.27& -14.19& -19.1 & -16.69\\
6  & -16.17& -16.02& -15.22& -16.19& -21.11& -18.69\\
7  & -18.17& -18.02& -17.21& -18.19& -23.11& -20.69\\
8  & -20.17& -20.02& -19.21& -20.19& -25.11& -22.69\\
\bottomrule
\end{tabular}
\normalsize
\label{table:classRelDiffMeans32}
\end{table}

\begin{table}[h!]
\caption{Mean rel.\ difference $\log_2(|\text{CKA}_{G(\sigma)}-\text{CKA}_{\text{lin}}|/\text{CKA}_{\text{lin}})$. \\NN width $w=64$. Classification.}	% 50 reps
\centering
\begin{tabular}{ccccccc}
\toprule
$\log_2 \sigma$ &splice &tic-tac-toe &wdbc &optdigits &wine &dna\\
\midrule
-4  &  -1.57&  -0.99&  -4.30&  -3.18&  -3.21&  -3.43\\
       -3  &  -3.44&  -1.18&  -5.4 &  -1.78&  -4.15&  -1.32\\
       -2  &  -0.86&  -4.28&  -6.69&  -4.70&  -6.03&  -0.53\\
       -1  &  -3.44&  -5.30&  -7.89&  -4.81&  -8.24&  -3.42\\
        0  &  -7.45&  -5.12&  -9.74&  -5.67& -10.49&  -6.29\\
        1  &  -8.09&  -6.32& -11.90&  -7.25& -13.27&  -9.53\\
        2  &  -9.34&  -8.12& -11.71&  -9.14& -15.92& -11.70\\
        3  & -11.19& -10.06& -11.51& -11.12& -18.06& -13.91\\
        4  & -13.15& -12.05& -12.87& -13.11& -20.29& -15.93\\
        5  & -15.14& -14.05& -14.71& -15.11& -22.31& -17.96\\
        6  & -17.14& -16.05& -16.67& -17.11& -24.24& -19.96\\
        7  & -19.14& -18.05& -18.66& -19.11& -26.23& -21.97\\
        8  & -21.14& -20.05& -20.66& -21.11& -28.23& -23.97\\
\bottomrule
\end{tabular}
\normalsize
\label{table:classRelDiffMeans64}
\end{table}

\begin{table}[h!]
\caption{Mean rel.\ difference $\log_2(|\text{CKA}_{G(\sigma)}-\text{CKA}_{\text{lin}}|/\text{CKA}_{\text{lin}})$. \\NN width $w=128$. Classification.}	% 50 reps
\centering
\begin{tabular}{ccccccc}
\toprule
$\log_2 \sigma$ &splice &tic-tac-toe &wdbc &optdigits &wine &dna\\
\midrule
-4  &  -1.53&  -0.97&  -5.84&  -2.96&  -3.82&  -2.76\\
       -3  &  -1.75&  -1.04&  -7.05&  -3.04&  -5.25&  -3.80\\
       -2  &  -1.03&  -2.83&  -8.33&  -5.01&  -7.65&  -0.46\\
       -1  &  -3.31&  -4.17&  -9.70&  -5.24& -10.18&  -3.13\\
        0  &  -7.09&  -4.49& -11.86&  -6.08& -12.56&  -5.65\\
        1  &  -9.31&  -5.94& -13.17&  -7.72& -15.33&  -8.56\\
        2  & -10.32&  -7.80& -13.25&  -9.63& -18.46& -11.12\\
        3  & -12.10&  -9.76& -13.76& -11.61& -20.20& -13.27\\
        4  & -14.05& -11.75& -15.07& -13.60& -22.23& -15.45\\
        5  & -16.04& -13.75& -16.90& -15.60& -24.26& -17.33\\
        6  & -18.04& -15.75& -18.86& -17.60& -26.27& -19.32\\
        7  & -20.04& -17.75& -20.85& -19.60& -28.27& -21.32\\
        8  & -22.04& -19.75& -22.85& -21.60& -30.27& -23.32\\
\bottomrule
\end{tabular}
\normalsize
\label{table:classRelDiffMeans128}
\end{table}

\begin{table}[h!]
\caption{Mean rel.\ difference $\log_2(|\text{CKA}_{G(\sigma)}-\text{CKA}_{\text{lin}}|/\text{CKA}_{\text{lin}})$. \\NN width $w=256$. Classification.}	% 50 reps
\centering
\begin{tabular}{ccccccc}
\toprule
$\log_2 \sigma$ &splice &tic-tac-toe &wdbc &optdigits &wine &dna\\
\midrule
-4  &  -1.43&  -0.94&  -6.75&  -3.12&  -4.30&  -2.52\\
       -3  &  -1.48&  -0.97&  -7.98&  -4.65&  -5.93&  -3.99\\
       -2  &  -1.31&  -2.14&  -9.24&  -5.54&  -8.42&  -0.47\\
       -1  &  -3.22&  -3.25& -10.63&  -5.35& -10.98&  -2.88\\
        0  &  -6.27&  -4.05& -12.57&  -6.28& -13.40&  -5.30\\
        1  & -10.21&  -5.59& -14.61&  -7.94& -16.01&  -7.96\\
        2  & -12.02&  -7.46& -14.16&  -9.85& -18.90& -10.36\\
        3  & -13.77&  -9.43& -14.44& -11.83& -21.26& -12.50\\
        4  & -15.70& -11.42& -15.80& -13.82& -23.34& -14.52\\
        5  & -17.69& -13.42& -17.64& -15.82& -25.34& -16.52\\
        6  & -19.67& -15.42& -19.61& -17.82& -27.34& -18.52\\
        7  & -21.66& -17.42& -21.60& -19.82& -29.33& -20.52\\
        8  & -23.66& -19.42& -23.59& -21.82& -31.33& -22.53\\
\bottomrule
\end{tabular}
\normalsize
\label{table:classRelDiffMeans256}
\end{table}

\begin{table}[h!]
\caption{Mean rel.\ difference $\log_2(|\text{CKA}_{G(\sigma)}-\text{CKA}_{\text{lin}}|/\text{CKA}_{\text{lin}})$. \\NN width $w=512$. Classification.}	% 50 reps
\centering
\begin{tabular}{ccccccc}
\toprule
$\log_2 \sigma$ &splice &tic-tac-toe &wdbc &optdigits &wine &dna\\
\midrule
-4 & -1.34 & -0.60 & -7.68 & -3.27 & -5.05 & -2.40 \\
       -3 & -1.39 & -0.63 & -8.90 & -5.91 & -7.02 & -3.27 \\
       -2 & -1.16 & -1.71 &-10.17 & -5.70 & -9.59 & -0.45 \\
       -1 & -3.01 & -2.81 &-11.52 & -5.51 &-12.10 & -2.68 \\
        0 & -5.71 & -3.69 &-13.38 & -6.48 &-14.47 & -5.09 \\
        1 & -9.51 & -5.23 &-14.60 & -8.15 &-17.21 & -7.63 \\
        2 &-12.15 & -7.10 &-14.53 &-10.06 &-19.76 & -9.91 \\
        3 &-14.51 & -9.06 &-14.86 &-12.04 &-21.92 &-12.00 \\
        4 &-16.45 &-11.05 &-16.14 &-14.04 &-24.06 &-14.02 \\
        5 &-18.45 &-13.05 &-17.98 &-16.03 &-26.11 &-16.02 \\
        6 &-20.45 &-15.05 &-19.94 &-18.03 &-28.13 &-18.03 \\
        7 &-22.45 &-17.05 &-21.93 &-20.03 &-30.13 &-20.03 \\
        8 &-24.45 &-19.05 &-23.93 &-22.03 &-32.13 &-22.03 \\
\bottomrule
\end{tabular}
\normalsize
\label{table:classRelDiffMeans512}
\end{table}

\begin{table}[h!]
\caption{Mean rel.\ difference $\log_2(|\text{CKA}_{G(\sigma)}-\text{CKA}_{\text{lin}}|/\text{CKA}_{\text{lin}})$. \\NN width $w=1024$. Classification.}	% 50 reps
\centering
\begin{tabular}{ccccccc}
\toprule
$\log_2 \sigma$ &splice &tic-tac-toe &wdbc &optdigits &wine &dna\\
\midrule
		-4 & -1.40 & -0.67 & -7.96 & -3.29 & -5.01 & -2.44\\
       -3 & -1.57 & -0.72 & -9.40 & -4.01 & -6.95 & -4.48\\
       -2 & -0.72 & -1.97 &-10.78 & -4.81 & -9.55 & -0.38\\
       -1 & -2.73 & -3.06 &-11.84 & -5.81 &-12.18 & -2.56\\
        0 & -5.28 & -3.93 &-13.24 & -6.60 &-14.67 & -4.98\\
        1 & -8.57 & -5.46 &-13.84 & -8.23 &-17.52 & -7.49\\
        2 &-11.26 & -7.32 &-13.95 &-10.13 &-20.21 & -9.75\\
        3 &-13.32 & -9.29 &-14.43 &-12.11 &-22.60 &-11.83\\
        4 &-15.36 &-11.28 &-15.89 &-14.10 &-24.46 &-13.85\\
        5 &-17.37 &-13.28 &-17.73 &-16.10 &-26.47 &-15.86\\
        6 &-19.37 &-15.28 &-19.70 &-18.10 &-28.48 &-17.86\\
        7 &-21.37 &-17.28 &-21.69 &-20.10 &-30.48 &-19.86\\
        8 &-23.37 &-19.28 &-23.69 &-22.10 &-32.49 &-21.86\\
\bottomrule
\end{tabular}
\normalsize
\label{table:classRelDiffMeans1024}
\end{table}

\begin{table}[h!]
\caption{Mean rel.\ difference $\log_2(|\text{CKA}_{G(\sigma)}-\text{CKA}_{\text{lin}}|/\text{CKA}_{\text{lin}})$. \\Three-layer 128-32-8 NN. Classification.}	% 50 reps
\centering
\begin{tabular}{ccccccc}
\toprule
$\log_2 \sigma$ &splice &tic-tac-toe &wdbc &optdigits &wine &dna\\
\midrule
 		-4 & -2.46 &  2.82 & -3.03 & -2.1  & -1.82 & -1.13\\
       -3 & -0.68 &  1.76 & -4.02 & -0.76 & -1.91 & -0.36\\
       -2 & -0.38 &  0.50 & -5.2  & -4.57 & -2.72 & -0.29\\
       -1 & -2.94 & -0.03 & -6.63 & -2.64 & -4.34 & -3.21\\
        0 & -6.07 & -1.20 & -8.5  & -3.76 & -6.33 & -5.92\\
        1 & -8.57 & -2.64 & -9.48 & -5.46 & -8.96 & -8.69\\
        2 & -9.72 & -4.40 & -9.24 & -7.38 &-11.05 &-11.03\\
        3 &-11.57 & -6.32 & -9.68 & -9.36 &-13.27 &-13.14\\
        4 &-13.49 & -8.31 &-11.03 &-11.35 &-15.29 &-15.03\\
        5 &-15.47 &-10.30 &-12.87 &-13.35 &-17.26 &-17.01\\
        6 &-17.47 &-12.30 &-14.85 &-15.35 &-19.26 &-19.00\\
        7 &-19.47 &-14.30 &-16.85 &-17.35 &-21.26 &-21.00\\
        8 &-21.46 &-16.30 &-18.85 &-19.35 &-23.26 &-23.00\\
\bottomrule
\end{tabular}
\normalsize
\label{table:classRelDiffMeansThreeHiddenLayers}
\end{table}

\begin{table}[h!]
\caption{Mean rel.\ difference $\log_2(|\text{CKA}_{G(\sigma)}-\text{CKA}_{\text{lin}}|/\text{CKA}_{\text{lin}})$. \\NN width $w=16$. Regression.}	% 50 reps
\centering
\begin{tabular}{ccccccc}
\toprule
$\log_2 \sigma$ &cpu &boston &Diab(skl) &stock &balloon &cloud\\
\midrule
	-4 &-3.28 &-2.15 &-2.11 &-2.76 &-8.00 &-3.19\\
    -3 &-3.38 &-2.02 &-3.95 &-2.53 &-8.26 &-2.38\\
    -2 &-4.06 &-2.56 &-6.42 &-2.25 &-8.25 &-2.16\\
    -1 &-5.25 &-3.49 &-9.22 &-2.95 &-8.58 &-2.75\\
     0 &-6.73 &-4.42 &-11.72 &-4.42 &-9.08 &-3.70 \\
     1 &-8.14 &-4.95 &-14.04 &-5.90  &-9.26 &-4.90 \\
     2 &-9.67 &-6.09 &-16.74 &-7.77 &-9.63 &-6.70 \\
     3 &-9.78 &-7.81 &-19.29 &-9.73 &-10.23 &-8.85\\
     4 &-10.63 &-9.77 &-21.56 &-11.71 &-11.50  &-11.03\\
     5 &-12.57 &-11.76 &-23.69 &-13.70  &-12.94 &-13.16\\
     6 &-14.42 &-13.76 &-25.70  &-15.70  &-14.81 &-15.23\\
     7 &-16.40  &-15.76 &-27.70  &-17.70  &-16.75 &-17.22\\
     8 &-18.39 &-17.76 &-29.70  &-19.70  &-18.73 &-19.22\\
\bottomrule
\end{tabular}
\normalsize
\label{table:regrRelDiffMeans}
\end{table}

\begin{table}[h!]
\caption{Mean rel.\ difference $\log_2(|\text{CKA}_{G(\sigma)}-\text{CKA}_{\text{lin}}|/\text{CKA}_{\text{lin}})$. \\NN width $w=32$. Regression.}	% 50 reps
\centering
\begin{tabular}{ccccccc}
\toprule
$\log_2 \sigma$ &cpu &boston &Diab(skl) &stock &balloon &cloud\\
\midrule
-4  &  -4.60 &  -2.91&  -1.63&  -2.90 &  -7.09&  -3.94\\
       -3  &  -4.67&  -2.84&  -3.43&  -2.89&  -7.61&  -4.59\\
       -2  &  -5.43&  -3.57&  -5.88&  -2.47&  -7.47&  -3.74\\
       -1  &  -6.77&  -4.74&  -8.64&  -2.82&  -7.50&  -3.62\\
        0  &  -7.95&  -4.92& -11.10&  -4.25&  -7.71&  -4.48\\
        1  &  -9.01&  -5.67& -13.36&  -6.00&  -8.17&  -5.53\\
        2  & -10.07&  -7.08& -16.02&  -7.78&  -8.49&  -7.37\\
        3  & -10.37&  -8.95& -18.57&  -9.75&  -9.84&  -9.76\\
        4  & -11.42& -10.88& -20.81& -11.77& -10.58& -11.83\\
        5  & -13.24& -12.87& -22.88& -13.80& -12.05& -13.88\\
        6  & -15.21& -14.87& -24.90& -15.82& -13.95& -15.88\\
        7  & -17.20& -16.87& -26.90& -17.83& -15.99& -17.88\\
        8  & -19.20& -18.87& -28.90& -19.84& -17.94& -19.88\\
\bottomrule
\end{tabular}
\normalsize
\label{table:regrRelDiffMeans32}
\end{table}

\begin{table}[h!]
\caption{Mean rel.\ difference $\log_2(|\text{CKA}_{G(\sigma)}-\text{CKA}_{\text{lin}}|/\text{CKA}_{\text{lin}})$. \\NN width $w=64$. Regression.}	% 50 reps
\centering
\begin{tabular}{ccccccc}
\toprule
$\log_2 \sigma$ &cpu &boston &Diab(skl) &stock &balloon &cloud\\
\midrule
-4  &  -6.48&  -3.74&  -1.34&  -3.27&  -6.41&  -3.78\\
       -3  &  -6.06&  -3.94&  -3.23&  -3.45&  -6.85&  -4.63\\
       -2  &  -6.71&  -4.30&  -5.85&  -2.95&  -6.70&  -5.10 \\
       -1  &  -7.69&  -5.24&  -8.71&  -3.56&  -6.62&  -4.85\\
        0  &  -9.22&  -5.40& -11.15&  -4.83&  -6.66&  -5.72\\
        1  & -10.64&  -5.81& -13.42&  -6.37&  -6.96&  -6.34\\
        2  & -11.59&  -7.36& -16.22&  -8.22&  -7.81&  -7.71\\
        3  & -12.25&  -9.30& -18.91& -10.14&  -9.37& -10.15\\
        4  & -13.37& -11.34& -21.14& -12.13& -10.42& -12.16\\
        5  & -15.17& -13.41& -23.21& -14.12& -11.99& -14.17\\
        6  & -17.09& -15.40& -25.23& -16.12& -13.92& -16.17\\
        7  & -19.08& -17.39& -27.24& -18.12& -15.91& -18.16\\
        8  & -21.07& -19.39& -29.24& -20.12& -17.90& -20.16\\
\bottomrule
\end{tabular}
\normalsize
\label{table:regrRelDiffMeans64}
\end{table}

\begin{table}[h!]
\caption{Mean rel.\ difference $\log_2(|\text{CKA}_{G(\sigma)}-\text{CKA}_{\text{lin}}|/\text{CKA}_{\text{lin}})$. \\NN width $w=128$. Regression.}	% 50 reps
\centering
\begin{tabular}{ccccccc}
\toprule
$\log_2 \sigma$ &cpu &boston &Diab(skl) &stock &balloon &cloud\\
\midrule
-4  &  -7.58&  -4.12&  -1.10&  -3.42&  -5.11&  -3.80\\
       -3  &  -7.11&  -4.02&  -2.94&  -3.51&  -5.31&  -4.32\\
       -2  &  -7.86&  -4.92&  -5.60&  -3.11&  -5.26&  -5.37\\
       -1  &  -8.80&  -6.51&  -8.49&  -3.65&  -5.23&  -5.38\\
        0  & -10.61&  -7.14& -10.94&  -4.65&  -5.26&  -6.05\\
        1  & -11.41&  -7.45& -13.18&  -6.09&  -5.46&  -6.85\\
        2  & -12.66&  -9.00& -16.06&  -7.92&  -5.96&  -8.36\\
        3  & -12.60& -10.99& -18.78&  -9.88&  -7.47& -10.43\\
        4  & -13.70& -13.04& -21.18& -11.86&  -8.63& -12.50\\
        5  & -15.50& -14.99& -23.36& -13.86& -10.28& -14.45\\
        6  & -17.47& -16.98& -25.38& -15.86& -12.22& -16.42\\
        7  & -19.46& -18.98& -27.37& -17.86& -14.21& -18.41\\
        8  & -21.46& -20.98& -29.37& -19.86& -16.20& -20.41\\
\bottomrule
\end{tabular}
\normalsize
\label{table:regrRelDiffMeans128}
\end{table}

\begin{table*}[h!]
\caption{Median ratios, $\rho = \max(\text{diam}(X)/d_X, \text{diam}(Y)/d_Y)$, with $\pm 2$ standard error equivalent confidence intervals,\\of maximum to median distance between features. $w$ is NN width. Classification.}
% 50 reps
\centering
\footnotesize
\begin{tabular}{ccccccc}
\toprule
$w$ &splice &tic-tac-toe &wdbc &optdigits &wine &dna\\
\midrule
%$\rho \pm 2\frac{\text{IQR}}{\sqrt{50}}$
16 &3.91 $\pm$ 0.21 &3.53 $\pm$ 0.30 &10.88 $\pm$ 0.56 &3.33 $\pm$ 0.29 &5.00 $\pm$ 0.01 &2.87 $\pm$ 0.08\\
32 &3.67 $\pm$ 0.13 &2.71 $\pm$ 0.14 &10.82 $\pm$ 0.26 &2.42 $\pm$ 0.08 &5.00 $\pm$ 0.01 &2.68 $\pm$ 0.06 \\
64 &3.14 $\pm$ 0.10 &2.43 $\pm$ 0.10 &10.74 $\pm$ 0.18 &2.15 $\pm$ 0.04 &5.01 $\pm$ 0.00 &2.55 $\pm$ 0.04 \\
128 &2.95 $\pm$ 0.07 &2.36 $\pm$ 0.05 &10.64 $\pm$ 0.09 &2.00 $\pm$ 0.05 &5.00 $\pm$ 0.00 &2.47 $\pm$ 0.03\\
256 &2.83 $\pm$ 0.06 &2.34 $\pm$ 0.07 &10.57 $\pm$ 0.06 &1.92 $\pm$ 0.03 &5.00 $\pm$ 0.00 &2.43 $\pm$ 0.02\\
512 &2.76 $\pm$ 0.05 &2.48 $\pm$ 0.06 &10.57 $\pm$ 0.06 &1.87 $\pm$ 0.03 &5.00 $\pm$ 0.00 &2.39 $\pm$ 0.02\\
1024 &2.74 $\pm$ 0.05 &2.45 $\pm$ 0.04 &10.62 $\pm$ 0.06 &1.89 $\pm$ 0.03 &5.01 $\pm$ 0.00 &2.39 $\pm$ 0.02\\
\midrule
128-32-8	&3.28 $\pm$ 0.12 &4.23 $\pm$ 0.33 &10.87 $\pm$ 0.26 &2.34 $\pm$ 0.08 &5.00 $\pm$ 0.00 &2.55 $\pm$ 0.08\\
\bottomrule
\end{tabular}
%\normalsize
\label{table:classRhoMeansIQR}
\end{table*}

\begin{table}[h!]
\caption{Mean rel.\ difference $\log_2(|\text{CKA}_{G(\sigma)}-\text{CKA}_{\text{lin}}|/\text{CKA}_{\text{lin}})$. \\NN width $w=256$. Regression.}	% 50 reps
\centering
\begin{tabular}{ccccccc}
\toprule
$\log_2 \sigma$ &cpu &boston &Diab(skl) &stock &balloon &cloud\\
\midrule
-4  &  -8.24&  -4.48&  -0.90&  -2.70&  -4.23&  -3.26\\
       -3  &  -7.87&  -4.47&  -2.61&  -2.74&  -4.35&  -3.56\\
       -2  &  -8.71&  -5.97&  -5.28&  -2.45&  -4.33&  -5.20\\
       -1  &  -9.64&  -7.36&  -8.22&  -2.94&  -4.31&  -6.21\\
        0  & -11.44&  -7.40& -10.70&  -4.17&  -4.32&  -6.63\\
        1  & -12.89&  -7.67& -13.00&  -5.66&  -4.44&  -7.99\\
        2  & -13.74&  -9.13& -15.68&  -7.51&  -4.81&  -8.36\\
        3  & -14.18& -10.99& -18.37&  -9.46&  -6.17& -10.52\\
        4  & -15.26& -12.96& -20.66& -11.45&  -7.44& -12.68\\
        5  & -17.03& -14.95& -22.75& -13.45&  -9.08& -14.69\\
        6  & -18.97& -16.95& -24.78& -15.45& -11.00& -16.69\\
        7  & -20.95& -18.95& -26.79& -17.45& -12.98& -18.69\\
        8  & -22.95& -20.95& -28.79& -19.45& -14.98& -20.69\\
\bottomrule
\end{tabular}
\normalsize
\label{table:regrRelDiffMeans256}
\end{table}

\begin{table}[h!]
\caption{Mean rel.\ difference $\log_2(|\text{CKA}_{G(\sigma)}-\text{CKA}_{\text{lin}}|/\text{CKA}_{\text{lin}})$. \\NN width $w=512$. Regression.}	% 50 reps
\centering
\begin{tabular}{ccccccc}
\toprule
$\log_2 \sigma$ &cpu &boston &Diab(skl) &stock &balloon &cloud\\
\midrule
	   -4 & -8.29 & -3.78 & -0.75 & -2.35 & -3.68 & -2.91\\
       -3 & -8.12 & -3.80 & -2.10 & -2.39 & -3.81 & -3.25\\
       -2 & -9.00 & -5.10 & -4.40 & -2.20 & -3.78 & -5.28\\
       -1 & -9.90 & -7.37 & -7.15 & -2.64 & -3.76 & -6.29\\
        0 &-11.48 & -7.91 & -9.61 & -3.78 & -3.79 & -6.54\\
        1 &-13.07 & -8.15 &-11.85 & -5.31 & -3.91 & -7.45\\
        2 &-13.93 & -9.65 &-14.89 & -7.17 & -4.24 & -7.95\\
        3 &-15.26 &-11.56 &-18.07 & -9.13 & -5.56 & -9.79\\
        4 &-16.44 &-13.51 &-20.29 &-11.12 & -6.86 &-11.83\\
        5 &-18.24 &-15.49 &-22.44 &-13.12 & -8.35 &-13.85\\
        6 &-20.22 &-17.49 &-24.47 &-15.12 &-10.25 &-15.86\\
        7 &-22.23 &-19.49 &-26.52 &-17.12 &-12.22 &-17.86\\
        8 &-24.23 &-21.49 &-28.50 &-19.12 &-14.22 &-19.86\\
\bottomrule
\end{tabular}
\normalsize
\label{table:regrRelDiffMeans512}
\end{table}

\begin{table}[h!]
\caption{Mean rel.\ difference $\log_2(|\text{CKA}_{G(\sigma)}-\text{CKA}_{\text{lin}}|/\text{CKA}_{\text{lin}})$. \\NN width $w=1024$. Regression.}	% 50 reps
\centering
\begin{tabular}{ccccccc}
\toprule
$\log_2 \sigma$ &cpu &boston &Diab(skl) &stock &balloon &cloud\\
\midrule
	   -4 & -6.87 & -3.35 & -0.67 & -2.18 & -3.26 & -2.72\\
       -3 & -6.86 & -3.19 & -1.71 & -2.22 & -3.39 & -3.11\\
       -2 & -7.90 & -4.01 & -3.64 & -2.05 & -3.35 & -5.58\\
       -1 & -8.70 & -6.49 & -5.97 & -2.46 & -3.33 & -5.28\\
        0 &-10.07 & -7.61 & -8.20 & -3.61 & -3.36 & -6.10\\
        1 &-11.85 & -7.65 &-10.35 & -5.15 & -3.48 & -7.13\\
        2 &-13.22 & -8.89 &-13.34 & -7.01 & -3.78 & -7.77\\
        3 &-15.36 &-10.72 &-16.72 & -8.98 & -4.91 & -9.62\\
        4 &-17.14 &-12.68 &-19.23 &-10.97 & -6.16 &-11.53\\
        5 &-18.91 &-14.67 &-21.48 &-12.97 & -7.69 &-13.53\\
        6 &-20.81 &-16.66 &-23.63 &-14.97 & -9.58 &-15.53\\
        7 &-22.80 &-18.66 &-25.64 &-16.97 &-11.55 &-17.53\\
        8 &-24.80 &-20.66 &-27.63 &-18.97 &-13.55 &-19.53\\
\bottomrule
\end{tabular}
\normalsize
\label{table:regrRelDiffMeans1024}
\end{table}

\begin{table}[h!]
\caption{Mean rel.\ difference $\log_2(|\text{CKA}_{G(\sigma)}-\text{CKA}_{\text{lin}}|/\text{CKA}_{\text{lin}})$. \\Three-layer 128-32-8 NN. Regression.}	% 50 reps
\centering
\begin{tabular}{ccccccc}
\toprule
$\log_2 \sigma$ &cpu &boston &Diab(skl) &stock &balloon &cloud\\
\midrule
-4 & -4.25 & -2.12 & -0.78 & -1.29 & -3.81 & -3.40 \\
       -3 & -4.20 & -1.87 & -2.17 & -0.79 & -4.05 & -2.85\\
       -2 & -4.91 & -2.27 & -4.50 & -0.31 & -4.06 & -2.29\\
       -1 & -5.96 & -3.41 & -7.20 & -0.67 & -3.99 & -2.75\\
        0 & -7.90 & -4.93 & -9.61 & -2.32 & -4.10 & -4.03\\
        1 & -9.61 & -5.34 &-11.94 & -4.36 & -4.36 & -5.54\\
        2 &-10.09 & -6.66 &-14.60 & -6.31 & -5.03 & -6.30\\
        3 &-10.43 & -8.49 &-17.68 & -8.28 & -6.84 & -8.70\\
        4 &-11.44 &-10.43 &-19.71 &-10.27 & -7.76 &-10.94\\
        5 &-13.27 &-12.41 &-21.68 &-12.27 & -9.48 &-12.94\\
        6 &-15.25 &-14.41 &-23.69 &-14.27 &-11.16 &-14.90\\
        7 &-17.25 &-16.41 &-25.69 &-16.27 &-13.06 &-16.90\\
        8 &-19.25 &-18.41 &-27.69 &-18.27 &-15.04 &-18.90\\
\bottomrule
\end{tabular}
\normalsize
\label{table:regrRelDiffMeansThreeHiddenLayers}
\end{table}

%} %Appendix / supplementary material

\end{document}